\documentclass{article}
\PassOptionsToPackage{table}{xcolor}
\PassOptionsToPackage{hypertexnames=false,hyperfootnotes=false}{hyperref}
\usepackage{colm2024_conference}
\usepackage[T1]{fontenc}
\usepackage{lmodern}

\usepackage{amsmath,amsfonts,bm}

\def\eqref#1{equation~\ref{#1}}

\def\1{\bm{1}}

\DeclareMathAlphabet{\mathsfit}{\encodingdefault}{\sfdefault}{m}{sl}
\SetMathAlphabet{\mathsfit}{bold}{\encodingdefault}{\sfdefault}{bx}{n}

\usepackage{hyperref}
\usepackage{url}
\usepackage{booktabs}
\usepackage{amsfonts}
\usepackage{amsmath}
\usepackage{amssymb}
\usepackage{microtype}
\usepackage{multirow}
\usepackage{xcolor}
\usepackage{array}
\usepackage{arydshln}
\usepackage{graphicx}
\usepackage{float}
\usepackage{flafter}
\usepackage{wrapfig}
\usepackage{placeins}
\usepackage{etoc}
\usepackage{tcolorbox}
\tcbuselibrary{skins,breakable,listings}
\usepackage[capitalise,noabbrev]{cleveref}
\usepackage{xspace}
\usepackage{marvosym}
\usepackage{listings}
\usepackage{pgfplots}
\pgfplotsset{compat=1.18}

\definecolor{posgreen}{RGB}{0,128,70}
\definecolor{negred}{RGB}{190,45,45}

\newcommand{\posdelta}[1]{\textcolor{posgreen}{#1}}
\newcommand{\negdelta}[1]{\textcolor{negred}{#1}}
\newcommand{\neudelta}[1]{\textcolor{gray}{#1}}

\title{\textsc{MetaExp}: Learning to Self-Improve through Experience via Verifiable Cross-Task Environment Scaling}

\author{
{Renhao Li}\textsuperscript{1,4,*},
{Xinjie Shen}\textsuperscript{2,*},
{Yang Su}\textsuperscript{3},
{Jianhong Tu}\textsuperscript{3,\dag}\\
{Derek F. Wong}\textsuperscript{1,\dag},
{Dayiheng Liu}\textsuperscript{3},
{Min Yang}\textsuperscript{4,\dag}\\[0.6em]
\textsuperscript{1}University of Macau,\quad
\textsuperscript{2}Georgia Institute of Technology\\
\textsuperscript{3}Alibaba Token Hub, Alibaba Group\\
\textsuperscript{4}Shenzhen Institute of Advanced Technology, Chinese Academy of Sciences\\
\;{\Letter}\; \url{li.renhao@connect.um.edu.mo}
}

\newcommand{\method}{\textsc{MetaExp}\xspace}
\newcommand{\protocol}{{SeqEvo}\xspace}

\begin{document}
\renewcommand{\thefootnote}{\fnsymbol{footnote}}
\maketitle
\footnotetext[1]{Equal contribution. Work done during the internship at Alibaba Token Hub.}
\footnotetext[2]{Corresponding author.}
\renewcommand{\thefootnote}{\arabic{footnote}}
\setcounter{footnote}{0}

\begin{abstract}

Language-model agents can accumulate interaction feedback, but reliably turning that feedback into reusable experience remains challenging. Existing methods either optimize external experience scaffolds or train experience-related behavior on specific task distributions, leaving open whether experience learning itself can become a transferable model capability. Training such a capability calls for supervision beyond isolated tasks, which mainly support in-episode problem solving rather than revealing what knowledge remains useful across episodes. In this work, we introduce \method, a scalable framework that trains this capability through verifiable task families with controlled cross-task structure. Each family preserves a reusable factor while varying task-specific realizations, enabling transfer without direct solution reuse. We propose two complementary designs: Shared-Tool families keep black-box tools fixed across tasks to elicit operational experience, while Shared-Workflow families preserve latent procedures across changing realizations to elicit procedural experience. We scale environments under both designs, collect and filter sequential interaction trajectories, and fine-tune Qwen3.5-122B-A10B on 4,956 trajectories from 1,000 task families. With parameters frozen during downstream evolution, \method improves by $+8.5$ points on SpreadsheetBench, peaks at $+2.3$ on APEX-Agents, and improves by $+14.8$ on AutomationBench. The gains persist under domain-excluded training, supporting experience learning as a transferable capability for continued non-parametric improvement after deployment.

\end{abstract}

\section{Introduction}

Large language models (LLMs) increasingly act in interactive environments, using tools to complete long-horizon tasks and respond to execution feedback~\citep{yao2023react}. Yet most agents remain largely episodic: solving one task does not by itself make the agent more effective on the next, and useful procedures discovered through interaction are often lost once the episode ends~\citep{zhao2024expel,wang2025agent}. Experience-driven self-evolution addresses this limitation by turning past trajectories and feedback into persistent experience that can guide future behavior~\citep{shinn2023reflexion}. However, retaining experience is not equivalent to learning from it. An effective agent must identify what caused an outcome, abstract what remains useful, update its experience accordingly, and reuse it in later tasks. This raises a fundamental question: can a model be trained once to acquire, update, and reuse experience more effectively on previously unseen task distributions?

Existing work approaches experience learning at either the system or model level. System-level methods keep the backbone fixed and improve how external experience is represented, extracted, revised, retrieved, or validated~\citep{ni2026trace2skill,yu2026skilladaptor,wang2026skillgrad,yang2026skillopt}. They demonstrate that external experience can support parameter-free adaptation, but its quality still depends on the pretrained model's ability to diagnose trajectories, abstract reusable knowledge, and apply it appropriately. Model-level approaches instead optimize experience acquisition, maintenance, or skill-conditioned execution through parameter training~\citep{wang-etal-2026-reinforcement,xia2026skillrl,fan2026evolving}. However, their optimization signals and downstream evaluation are typically tied to the same or closely related task distributions, making general experience-learning ability difficult to separate from task-specific adaptation. It therefore remains unclear whether experience learning itself can be trained as a transferable model capability.

This question is particularly relevant to realistic agent deployment. An agent may encounter a new task distribution without a curated skill library, large collections of verified target-domain trajectories, or the opportunity for benchmark-specific retraining. It must instead improve incrementally from the tasks and feedback it encounters, even when later task variants differ substantially from earlier ones. Motivated by this setting, we study \emph{meta experience learning}: after a single general training stage, the model parameters remain fixed, and the agent starts each held-out task distribution with an empty experience store. It must construct, revise, and reuse experience solely through sequential interaction and feedback. In this work, we instantiate external experience as structured Agent Skills~\citep{anthropic2025equipping}. This setting separates learning how to acquire and use experience from pre-acquiring or carrying target-domain experience itself.

Training this capability requires supervision beyond isolated task solving. Independent task trajectories primarily supervise in-episode problem solving, but provide limited evidence about which parts of an interaction will remain useful on later tasks. We therefore organize trajectory collection around structured task families that provide controlled cross-episode supervision. \method operationalizes this idea by constructing verifiable families under the principle of \emph{transfer without direct solution reuse}: each family preserves a reusable factor across tasks while varying task-specific realizations, so that experience acquired from earlier members can help with later ones without allowing previous solutions to be directly reused. We instantiate this principle through two complementary designs. \emph{Shared-Tool} families keep black-box tools fixed across tasks while varying their concrete problems, encouraging operational experience about recurring interfaces and tool composition. \emph{Shared-Workflow} families instead preserve a latent multi-step procedure across changing domains, tools, and data, encouraging higher-level procedural experience across different realizations. Sequential trajectories collected from environments scaled under both designs are then used to train the model to acquire, update, and reuse experience.

We fine-tune Qwen3.5-122B-A10B on 4,956 filtered trajectories collected from 1,000 synthetic task families and evaluate it under cold-start sequential evolution on three held-out agent benchmarks: SpreadsheetBench, APEX-Agents, and AutomationBench. As test-time experience accumulates, \method improves by $+8.5$ points on SpreadsheetBench, peaks at $+2.3$ points on APEX-Agents, and improves by $+14.8$ points on AutomationBench. In contrast, the matched Qwen3.5 backbone degrades by $2.2$ points on SpreadsheetBench under the same evolution protocol, showing that strong task-solving ability does not necessarily translate into effective experience learning. The gains remain substantial when semantically related training domains are excluded, and \method also makes better use of externally provided skills on SkillsBench. Together, these results support experience learning as a transferable model capability that can drive continued improvement through non-parametric experience accumulation on unseen task distributions.

Our contributions are threefold: (i) We formulate \emph{meta experience learning} as a transferable model capability: after a single general training stage, an agent should be able to build and reuse external experience from scratch on unseen task distributions without further parameter updates. (ii) We introduce \method, a scalable framework that constructs verifiable task families with controlled cross-task structure to generate cross-episode supervision for experience learning, instantiated through complementary Shared-Tool and Shared-Workflow designs. (iii) We demonstrate transferable cold-start experience learning across heterogeneous held-out agent benchmarks, with substantial gains retained under domain-excluded training and improved utilization of curated skills.

\section{Related Work}

\paragraph{Experience-driven agent self-evolution.}

Existing approaches improve agents through persistent experience at either the system or model level. System-level methods keep the backbone fixed and externalize experience as verbal memories, codes, or executable skills~\citep{shinn2023reflexion,zhao2024expel,wang2024voyager,wang2025agent,wang2025inducing}. Recent work further improves how such experience is extracted, revised, retrieved, and validated~\citep{ni2026trace2skill,qiu2026autorefine,jiang2026xskill,lin2026muse,yu2026skilladaptor,li2026mind,wang2026skillgrad,yang2026skillopt}. These approaches enable parameter-free adaptation, but the resulting experience remains dependent on the frozen model's ability to diagnose trajectories and abstract reusable knowledge, and some rely on substantial target-distribution trials or in-distribution validation~\citep{zhang2026coevoskills,liu2026skillrevise}. Model-level methods instead update parameters to optimize experience acquisition, maintenance, or skill-conditioned execution~\citep{lu2026skill0,zhu2026skill0,wu2025evolver,wang-etal-2026-reinforcement,xia2026skillrl,shi2026skill1,yang2026skillmaster,he2026reskill,zhang2026skillcomposer,yao2026skillrise,li2026codeskill,vishe2026skill,fan2026evolving}.
However, these methods are often trained and evaluated on the same or closely related task distributions, making transferable experience-learning ability difficult to assess. More recently, Connect the Dots (CoD)~\citep{chen2026connect} explicitly investigates cross-domain generalization of experience learning through reinforcement learning with credit assignment across task-solving and context-updating episodes. Its evaluation primarily focuses on controlled game-like and simulated science and engineering environments. In comparison, \method constructs verifiable task families with controlled cross-task structure to supervise operational and procedural experience learning from execution and reflection trajectories, and evaluates cold-start skill evolution across heterogeneous, work-oriented agent benchmarks.

\paragraph{Environment scaling and evaluation.}

Agent training data can be scaled through verifiable task and environment synthesis, including task proposal, capability composition, and automated construction of executable environments~\citep{zhou2025proposeragentevaluator,xie2026agentsynth,zhu2026termigen,hua2026cliuniverse,wei2026steps}. Some methods organize generation around predefined skills or skill structures~\citep{zheng2025skillweaver,fan2026skillsynth,cheng2026terminalworld,tan2026skt}. These approaches primarily scale individual tasks or environments, whereas \method scales \emph{task families} with controlled cross-task structure so that experience acquired from earlier tasks can remain useful for later ones without enabling direct solution reuse. Complementary benchmarks evaluate either the utility of provided skills~\citep{li2026skillsbench,han2026swe} or continual skill generation and refinement over task streams~\citep{zhong2026skilllearnbench,zhou2026skillgenbench,zhang2026skillflow,lei2026skillevolbench,jiang2026sea,gao2026evoagentbench,zhou2026gdpevo}. Our evaluation instead focuses on cold-start sequential evolution across heterogeneous held-out task distributions.

\section{Problem Setup}
\label{sec:formulation}

We study whether a language model can be trained once to better acquire, update, and reuse external experience on previously unseen task distributions. During test-time evolution, the model parameters remain fixed and only the accumulated experience changes. We first define this sequential experience-learning setting and then introduce the cold-start protocol used for evaluation.

\subsection{Experience Learning Setting}

We consider a sequence of tasks, where each task $\tau_i=(e_i,q_i,v_i)$ consists of an interactive environment $e_i$, a textual task request $q_i$, and an evaluator $v_i$. Given a model with parameters $\theta$ and an external experience store $B_{i-1}$, the agent interacts with the environment and produces a trajectory
\begin{equation}
    h_i \sim \pi_\theta(\cdot \mid e_i,q_i,B_{i-1}).
\end{equation}
The evaluator then returns an outcome score $r_i$ together with task-level feedback $f_i$,
\begin{equation}
    (r_i,f_i)=v_i(h_i).
\end{equation}
The score measures task completion, while the feedback contains richer information about successes and failures. Based on the trajectory and feedback, the agent may update its experience:
\begin{equation}
    B_i
    =
    U_\theta(B_{i-1},\tau_i,h_i,f_i),
\end{equation}
where $U_\theta$ denotes the model's experience-update behavior. The resulting $B_i$ persists across tasks and can influence subsequent execution. Experience may take different forms; in this work, we instantiate $B_i$ as a file-based bank of structured agent skills. Experience learning therefore involves both deciding what skills should be updated according to previous interactions and using the resulting skills appropriately on future tasks.

Our goal is not to learn target-domain experience during parameter training. Instead, we ask whether a single \emph{general training stage} can improve the model's ability to acquire and reuse experience under this sequential process on new task distributions. Training produces parameters $\theta_{\mathrm{ME}}$ from a base model $\theta_0$ using only source task families. At evaluation time, neither model receives target-benchmark parameter updates or pre-accumulated target-domain experience. Their parameters remain fixed, and adaptation occurs only through the evolving experience store $B$. We refer to this property as \emph{meta experience learning}: the ability to acquire useful experience from interaction feedback and reuse it on subsequent tasks, such that this behavior transfers beyond the task distributions used for parameter training. Operationally, it requires the model to construct experience from scratch, revise it as new feedback arrives, and selectively apply it when solving later tasks.

\subsection{Cold-Start Sequential Evolution}
\label{sec:protocol}

We introduce a cold-start sequential evolution protocol (\protocol) to evaluate experience learning on held-out task distributions. For a fixed evaluation domain, we omit the domain index for clarity. Both the adaptation and target tasks come from a task distribution that is not used for parameter optimization. The agent starts with an empty experience store,
\begin{equation}
    B_0=\varnothing,
\end{equation}
and processes $K$ adaptation tasks $(\tau_1,\ldots,\tau_K)$ one at a time. Following the update process above, the experience bank obtained after task $\tau_i$ is carried forward to $\tau_{i+1}$, yielding $B_k$ after the first $k$ tasks. No future adaptation task or target-set feedback is available when constructing $B_k$. Here, $k$ denotes the number of test-time adaptation tasks rather than gradient updates.

We evaluate each $B_k$ on a fixed target set $E$ that is disjoint from the adaptation tasks, while keeping the experience bank unchanged during evaluation. We define
\begin{equation}
    S(k;\theta)
    =
    \frac{1}{|E|}
    \sum_{\tau \in E}
    r_\tau^{(k,\theta)},
\end{equation}
where $r_\tau^{(k,\theta)}$ denotes the outcome score obtained by model $\theta$ on task $\tau$ using the accumulated experience bank $B_k$, and $S(0;\theta)$ is the performance without accumulated experience. The effect of test-time experience accumulation after $k$ adaptation tasks is
\begin{equation}
    \Delta(k;\theta)
    =
    S(k;\theta)-S(0;\theta).
\end{equation}
We report both $S(0;\theta)$ and $\Delta(k;\theta)$. The former reflects zero-experience task-solving ability, while the latter measures how performance changes as test-time experience accumulates. Together, they separate zero-experience task performance from the change induced by accumulated experience.

\begin{figure}[!t]
    \centering
    \includegraphics[width=\textwidth]{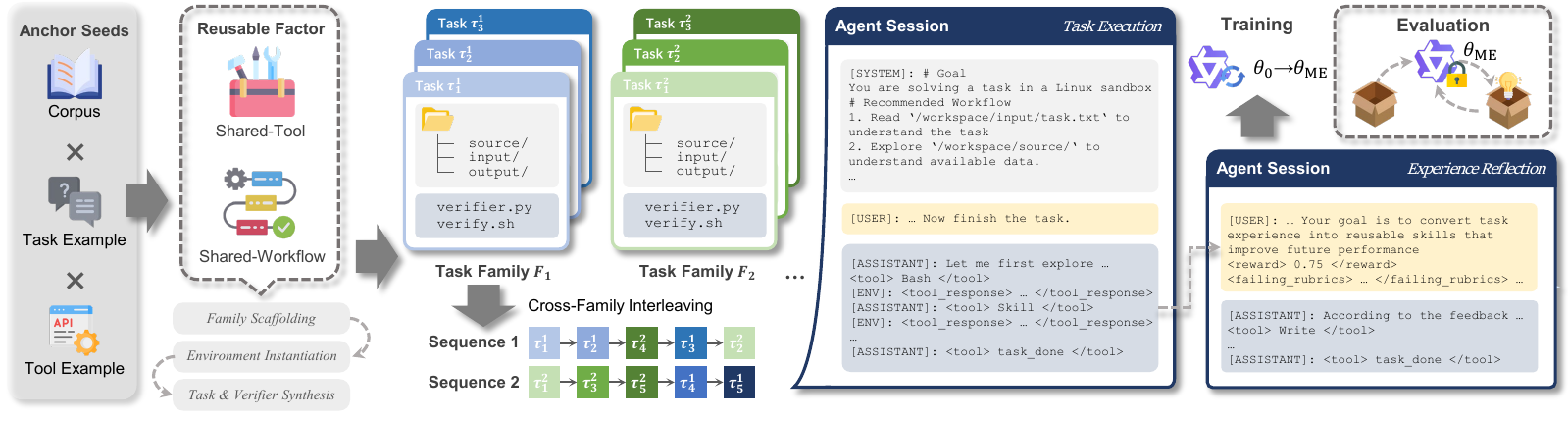}
    \caption{Overview of \method. Heterogeneous seeds are expanded into verifiable task families with complementary Shared-Tool and Shared-Workflow structure. Family-centered task sequences are used to collect execution and feedback-driven reflection trajectories, which train the model to acquire, update, and reuse experience. At evaluation, model parameters are frozen and experience is accumulated from an empty bank on unseen task distributions.}
    \label{fig:framework}
\end{figure}

\section{\method: Verifiable Cross-Task Environment Scaling}
\label{sec:method}

\method trains transferable experience-learning behavior by constructing controlled cross-task supervision. As shown in Figure~\ref{fig:framework}, it builds two complementary types of task families with reusable cross-task structure, realizes them as verifiable executable environments, collects sequential experience-learning trajectories, and uses the resulting data for model training.

\subsection{Cross-Task Family Design}

The key design principle is to preserve a reusable source of commonality across tasks while varying their concrete realizations. We represent a task family as
\begin{equation}
\mathcal{F}(z)=\{\tau_i=G(z,\xi_i)\}_{i=1}^{m},
\end{equation}
where $z$ denotes the family-level reusable factor and $\xi_i$ captures task-specific variation. We follow the principle of \emph{transfer without direct solution reuse}: $z$ should make experience acquired from earlier members potentially useful for later ones, while variation in $\xi_i$ prevents previous solutions from being directly reused.

We propose two complementary designs. (i) \textbf{Shared-Tool families} keep a black-box tool library and its documentation fixed across members while varying task data and requests. Here, $z$ corresponds to stable tool semantics and interfaces, encouraging relatively concrete operational experience about tool behavior, argument conventions, and composition. (ii) \textbf{Shared-Workflow families} instead preserve a latent multi-step procedure while varying its concrete realization, including domain, APIs and documentation, data, and request. The latent workflow guides generation but is never exposed to the solver, encouraging more abstract procedural experience that transfers across changing tools and domains. Together, the two designs expose reusable structure at complementary levels, from concrete operational regularities to abstract procedural patterns.

\subsection{Verifiable Family Scaling}

We scale these families from heterogeneous seeds drawn from pretraining-style corpora, task examples, and tool schemas, which provide diverse semantic contexts, procedural patterns, and executable affordances. Each family is then realized through three stages. \textbf{Family scaffolding} constructs its shared factor $z$ once per family. For Shared-Tool families, we generate and validate an executable tool library together with its documentation, which are then fixed across members. For Shared-Workflow families, we abstract a domain-independent latent workflow from a seed task example.

\textbf{Environment instantiation} produces multiple family members by varying task-specific factors $\xi_i$. Shared-Tool members reuse the same tools with independently populated data, whereas Shared-Workflow members instantiate the same latent procedure with new domains, APIs and documentation, and task-specific data. We additionally vary data scale, output structure, and difficulty to increase task diversity while preserving the intended commonality.

Finally, \textbf{task-and-verifier synthesis} uses a propose--solve--confirm loop to generate each task request together with a reference solution and executable checks. The generator explores the instantiated tools and data before proposing and solving the task. The complete package is rerun in a clean sandbox and retained only if the reference solution executes successfully and passes verification. Validation is asymmetric: reference solutions, verifiers, and latent workflows may be used during synthesis but are never exposed to the solver.

\subsection{Sequential Experience Trajectory Collection}

For each validated family $\mathcal{F}=\{\tau_1,\ldots,\tau_m\}$, we order its members by increasing difficulty and assign them sequentially to a teacher model starting from an empty skill bank. Each collection sequence is centered on one family but interleaves a small fraction of tasks sampled from other families. These cross-family tasks introduce potentially irrelevant experience into the bank, encouraging the model to learn selective reuse instead of indiscriminately applying accumulated skills. The teacher follows the sequential experience-learning process defined in Section~\ref{sec:formulation}: it solves each task conditioned on the current bank, receives executable feedback, and updates the bank before proceeding to the next task.

Each task runs in a fresh sandbox with the current skill bank mounted. The teacher interacts through a general agent harness and invokes a \texttt{task\_done} tool upon completion, after which the task-specific verifier is introduced and executed. An LLM judge reviews only failed rubric items to correct potential false negatives and produces the final score and feedback. The feedback is then injected through a fixed reflection prompt, allowing the teacher to create, revise, or delete skills before the updated bank is carried forward. We retain only trajectories with coherent execution, evaluation, and reflection, verified task completion, and valid skill artifacts.

\subsection{Meta-Experience Training}

The retained trajectories supervise two complementary behaviors: execution-stage actions train experience-conditioned task solving, while reflection-stage actions train feedback-driven experience updating. Let $\mathcal{D}_{\mathrm{ME}}$ denote records from both stages. We fine-tune the base model with the standard assistant-token objective
\begin{equation}
\mathcal{L}_{\mathrm{ME}}(\theta)
=
-\sum_{x\in\mathcal{D}_{\mathrm{ME}}}
\sum_{t\in\mathcal{A}(x)}
\log p_\theta(x_t\mid x_{<t}),
\end{equation}
where $\mathcal{A}(x)$ denotes assistant-generated token positions; system messages, user messages, and environment observations are masked from the loss.

Optimizing this objective from $\theta_0$ yields $\theta_{\mathrm{ME}}$. The skill banks accumulated during trajectory collection are discarded after model training; what is retained is the model's learned behavior for acquiring, updating, and reusing experience. At evaluation time, $\theta_{\mathrm{ME}}$ remains frozen and each held-out task distribution starts from an empty bank under \protocol, where adaptation occurs solely through newly accumulated experience.

\begin{table}[!t]
    \caption{
    Performance and efficiency on SpreadsheetBench. Values are mean $\pm$ population standard deviation, with $\Delta_k=S(k)-S(0)$ shown in parentheses. Green/red denotes improvement/degradation relative to $k=0$. EvoSkill and SkillGrad results are taken from \citet{wang2026skillgrad} and follow their originally reported evolution protocols; all other results use our cold-start \protocol protocol. ``--'' denotes an unreported checkpoint.
    }
    \label{tab:spreadsheet-main}
    \centering
    \small
    \setlength{\tabcolsep}{4.5pt}
    \renewcommand{\arraystretch}{1.15}
    \resizebox{\linewidth}{!}{
    \begin{tabular}{llccccc}

    \toprule

    \textbf{Model}
    & \textbf{Protocol}
    & $k=0$
    & $k=5$
    & $k=10$
    & $k=20$
    & $k=40$ \\

    \midrule


    \rowcolor{black!8}
    \multicolumn{7}{c}{\textit{Task success rate} (\%, $\uparrow$)} \\

    \addlinespace[2pt]

    GPT-4.1
    & EvoSkill
    & 44.2
    & --
    & --
    & --
    & $37.2\pm2.6$\,{\scriptsize\negdelta{$(-7.0)$}}
    \\

    GPT-5.4
    & EvoSkill
    & 62.5
    & --
    & --
    & --
    & $68.1\pm0.5$\,{\scriptsize\posdelta{$(+5.6)$}}
    \\

    GPT-4.1
    & SkillGrad
    & 44.2
    & --
    & --
    & --
    & $54.2\pm3.5$\,{\scriptsize\posdelta{$(+10.0)$}}
    \\

    GPT-5.4
    & SkillGrad
    & 62.5
    & --
    & --
    & --
    & $71.1\pm1.7$\,{\scriptsize\posdelta{$(+8.6)$}}
    \\

    \hdashline

    GLM-5.2
    & SeqEvo
    & $88.3\pm1.2$
    & $88.9\pm0.5$\,{\scriptsize\posdelta{$(+0.6)$}}
    & $89.5\pm0.7$\,{\scriptsize\posdelta{$(+1.2)$}}
    & $89.6\pm0.4$\,{\scriptsize\posdelta{$(+1.3)$}}
    & $89.4\pm0.4$\,{\scriptsize\posdelta{$(+1.1)$}}
    \\

    Qwen3.6-Plus
    & SeqEvo
    & $76.0\pm0.9$
    & $71.8\pm2.5$\,{\scriptsize\negdelta{$(-4.2)$}}
    & $72.9\pm3.1$\,{\scriptsize\negdelta{$(-3.1)$}}
    & $73.1\pm3.1$\,{\scriptsize\negdelta{$(-2.9)$}}
    & $74.3\pm3.1$\,{\scriptsize\negdelta{$(-1.7)$}}
    \\

    Qwen3.5-122B-A10B
    & SeqEvo
    & $68.9\pm1.8$
    & $67.3\pm1.8$\,{\scriptsize\negdelta{$(-1.6)$}}
    & $67.7\pm0.5$\,{\scriptsize\negdelta{$(-1.2)$}}
    & $67.0\pm0.9$\,{\scriptsize\negdelta{$(-1.9)$}}
    & $66.7\pm1.3$\,{\scriptsize\negdelta{$(-2.2)$}}
    \\

    \rowcolor{blue!10}
    \textbf{\textsc{\method}-Qwen3.5-122B-A10B}
    & SeqEvo
    & $65.0\pm2.2$
    & $69.5\pm3.3$\,{\scriptsize\bfseries\posdelta{$(+4.5)$}}
    & $72.9\pm3.7$\,{\scriptsize\bfseries\posdelta{$(+7.9)$}}
    & $73.1\pm2.8$\,{\scriptsize\bfseries\posdelta{$(+8.1)$}}
    & $73.5\pm3.4$\,{\scriptsize\bfseries\posdelta{$(+8.5)$}}
    \\

    \midrule


    \rowcolor{black!8}
    \multicolumn{7}{c}{\textit{Execution turns} ($\downarrow$)} \\

    \addlinespace[2pt]

    GLM-5.2
    & SeqEvo
    & $10.4\pm0.4$
    & $12.3\pm0.2$\,{\scriptsize\negdelta{$(+1.9)$}}
    & $12.1\pm0.1$\,{\scriptsize\negdelta{$(+1.7)$}}
    & $12.4\pm0.5$\,{\scriptsize\negdelta{$(+2.0)$}}
    & $13.1\pm0.6$\,{\scriptsize\negdelta{$(+2.7)$}}
    \\

    Qwen3.6-Plus
    & SeqEvo
    & $7.8\pm0.1$
    & $8.0\pm0.3$\,{\scriptsize\negdelta{$(+0.2)$}}
    & $8.0\pm0.2$\,{\scriptsize\negdelta{$(+0.2)$}}
    & $8.1\pm0.2$\,{\scriptsize\negdelta{$(+0.3)$}}
    & $7.7\pm0.2$\,{\scriptsize\posdelta{$(-0.1)$}}
    \\

    Qwen3.5-122B-A10B
    & SeqEvo
    & $7.6\pm0.5$
    & $7.2\pm0.1$\,{\scriptsize\posdelta{$(-0.4)$}}
    & $7.3\pm0.0$\,{\scriptsize\posdelta{$(-0.3)$}}
    & $7.4\pm0.1$\,{\scriptsize\posdelta{$(-0.2)$}}
    & $7.2\pm0.0$\,{\scriptsize\posdelta{$(-0.4)$}}
    \\

    \rowcolor{blue!10}
    \textbf{\textsc{\method}-Qwen3.5-122B-A10B}
    & SeqEvo
    & $8.1\pm0.1$
    & $8.1\pm0.2$\,{\scriptsize\neudelta{$(0.0)$}}
    & $8.2\pm0.3$\,{\scriptsize\negdelta{$(+0.1)$}}
    & $8.5\pm0.9$\,{\scriptsize\negdelta{$(+0.4)$}}
    & $8.6\pm0.7$\,{\scriptsize\negdelta{$(+0.5)$}}
    \\

    \bottomrule

    \end{tabular}
    }

\end{table}

\section{Experiments}
\label{sec:experiments}

\subsection{Experimental Setup}

\paragraph{Training configuration.}

Following Section~\ref{sec:method}, we construct 500 Shared-Tool and 500 Shared-Workflow families, each containing five task instances. We use GLM-5.2~\citep{zeng2026glm} as the teacher model, deployed through a standardized agent harness for trajectory collection, yielding 4,956 trajectories after filtering. We fine-tune Qwen3.5-122B-A10B~\citep{qwen2026qwen35} for two epochs with a 256K context window. The loss is computed only over assistant-generated tokens, while system messages, user messages, and environment observations are masked out.

\paragraph{Datasets and evaluation.}

We evaluate cold-start experience evolution on three challenging benchmarks spanning distinct task distributions. We adapt their original environments and evaluators into executable sandboxes compatible with our agent harness. All models evaluated by us follow the \protocol protocol defined in Section~\ref{sec:protocol}, using the same agent harness as in trajectory collection. For each benchmark, adaptation and target sets are disjoint, and we repeat each experiment three times with independently randomized adaptation orders. Each adaptation task allows up to three reflection retries; if all retries fail, the latest verified skill bank is carried forward. \textbf{SpreadsheetBench-v1-verified} contains 400 real-world spreadsheet manipulation tasks at the cell and sheet levels. We use 40 tasks for adaptation and 350 disjoint tasks for evaluation, with a shared skill bank capped at 10 skills. Verification executes the generated solution on the input workbook, recalculates formulas, and compares annotated output ranges with the golden workbook; a task succeeds only if all required cells match. \textbf{APEX-Agents} contains 480 long-horizon professional tasks across investment banking, law, and management consulting. For each domain, we use 40 tasks for adaptation and 120 disjoint tasks for evaluation, maintaining a separate skill bank capped at 5 skills. An LLM judge evaluates the agent response, generated artifacts, and environment changes against expert-authored binary rubrics; a task succeeds only if all required criteria are satisfied. \textbf{AutomationBench} contains 600 cross-application API workflow tasks across finance, HR, marketing, operations, sales, and support. For each domain, we use 20 tasks for adaptation and 80 disjoint tasks for evaluation, with an independent skill bank capped at 10 skills. Verification compares the final environment state against deterministic assertions. We report average task success rate on all three benchmarks. Each adaptation task is encountered only once per run, preventing repeated optimization against the same verifier feedback. We additionally evaluate the ability to utilize externally provided experience on \textbf{SkillsBench-v1.1}, which contains 87 expert-level tasks across 11 domains, each paired with curated agent skills. All models receive the same skills for each task, and we report average score and pass rate over three runs.

\paragraph{Models and baselines.}

We denote our trained checkpoint as \method-Qwen3.5-122B-A10B and use the original Qwen3.5-122B-A10B as the matched backbone baseline. We additionally evaluate stronger models, including Qwen3.6-Plus and GLM-5.2, with reasoning enabled at maximum effort. These models are evaluated using the same harness and \protocol setup. On SpreadsheetBench, we also report published EvoSkill~\citep{alzubi2026evoskill} and SkillGrad results from \citet{wang2026skillgrad} as representative system-level experience-learning approaches; these follow their originally reported evolution protocols rather than our \protocol setup.

\begin{figure}[!t]
    \centering
    \includegraphics[width=\textwidth]{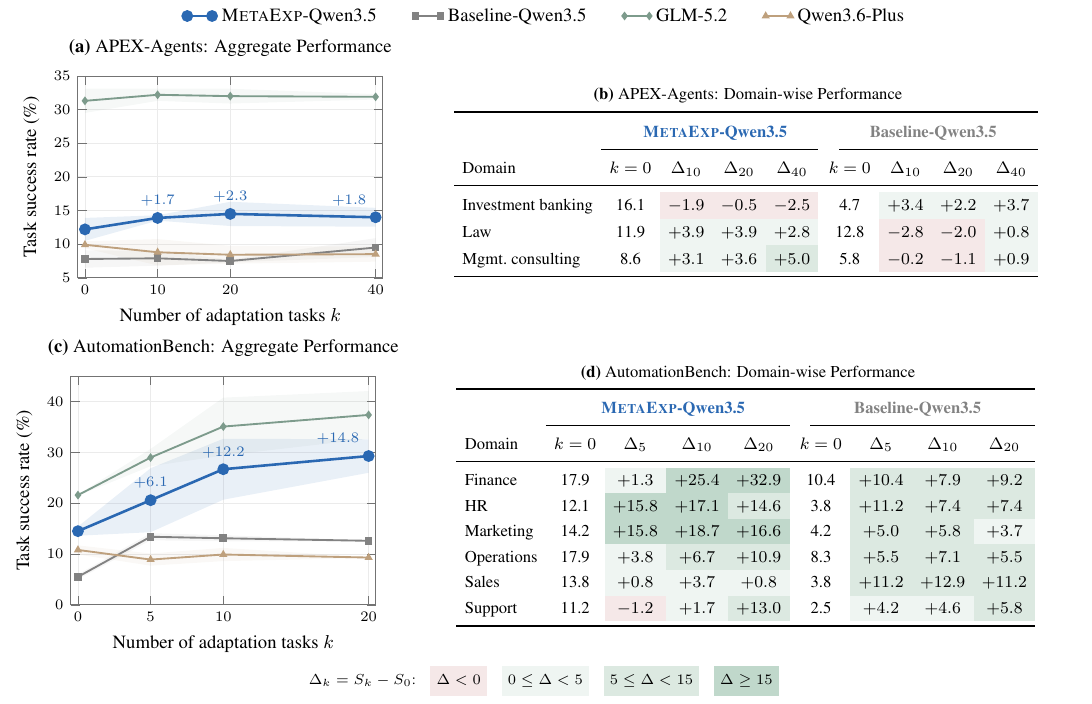}
    \caption{Performance on APEX-Agents and AutomationBench. Left: aggregate success rate over adaptation steps, with shaded population standard deviations over three runs. Right: domain-wise success rate and changes $\Delta_k$ on each adaptation step for \method-Qwen3.5-122B-A10B and Qwen3.5-122B-A10B; green/red denotes improvement/degradation.
    }
    \label{fig:apex-ab-main}
\end{figure}

\subsection{Main Results}

\paragraph{SpreadsheetBench.}

Table~\ref{tab:spreadsheet-main} reports performance and efficiency on SpreadsheetBench across different backbones and evolution protocols. Under \protocol, \method improves from $65.0$ at $k=0$ to $73.5$ at $k=40$ $(+8.5)$, with most gains emerging within the first 10 adaptation tasks. In contrast, Qwen3.5-122B-A10B and Qwen3.6-Plus finish below their zero-experience baselines, while GLM-5.2 remains the strongest absolute solver but improves only marginally, peaking at $+1.3$ points. The gain is accompanied by only a modest increase in execution turns, from $8.1$ at $k=0$ to $8.6$ at $k=40$, indicating that the improvement is not simply associated with substantially more interaction. Published EvoSkill and SkillGrad results further show that system-level skill evolution can also yield substantial gains, although they follow different evolution protocols and typically use more verified trajectories during adaptation.

\paragraph{APEX-Agents.}

Figure~\ref{fig:apex-ab-main}(a,b) shows a more heterogeneous pattern of experience transfer on APEX-Agents. At the aggregate level, \method improves from $12.2$ at $k=0$ to a peak of $14.5$ at $k=20$, retaining a $+1.8$-point gain at $k=40$. This modest average masks clear domain differences: at $k=40$, \method gains $+2.8$ points in law and $+5.0$ in management consulting, compared with only $+0.8$ and $+0.9$ for the matched Qwen3.5 baseline.

Investment banking is the main exception, where \method decreases by $2.5$ points while Qwen3.5 improves by $3.7$. This reversal is partly amplified by the near-binary exact-value evaluation and different cold-start headroom, although some case-specific skills also overgeneralize and interfere with otherwise correct solutions. GLM-5.2 remains the strongest absolute solver but shows little further improvement from accumulated experience. Overall, \method yields positive aggregate evolution, with particularly clear gains in law and management consulting.

\paragraph{AutomationBench.}

Figure~\ref{fig:apex-ab-main}(c,d) shows a broader and more sustained evolution trend on AutomationBench. \method improves monotonically from $14.5$ to $29.3$, yielding a $+14.8$-point gain, roughly twice that of its matched Qwen3.5 baseline. By $k=10$, \method already improves over its zero-experience performance in all six domains, and by $k=20$ its gain exceeds that of Qwen3.5 in five of them. The strongest improvement appears in finance ($+32.9$), with substantial gains also observed across HR, marketing, operations, and support. GLM-5.2 remains the strongest absolute solver and also benefits from accumulated experience. Overall, AutomationBench shows that \method substantially increases experience-driven improvement.

\begin{table}[!t]
    \caption{
    SpreadsheetBench task success under domain-excluded training.
    Values are mean $\pm$ population standard deviation over three runs, with $\Delta_k=S(k)-S(0)$ shown in parentheses.
    }
    \label{tab:strict-ood}
    \centering
    \small
    \setlength{\tabcolsep}{4.5pt}
    \renewcommand{\arraystretch}{1.15}
    \resizebox{\linewidth}{!}{
    \begin{tabular}{lccccc}
    \toprule
    \textbf{Model}
    & $k=0$
    & $k=5$
    & $k=10$
    & $k=20$
    & $k=40$ \\
    \midrule
    \rowcolor{black!8}
    \multicolumn{6}{c}{\textit{Task success rate} (\%, $\uparrow$)} \\
    \addlinespace[2pt]

    Qwen3.5-122B-A10B
    & $68.9\pm1.8$
    & $67.3\pm1.8$\,{\scriptsize\negdelta{$(-1.6)$}}
    & $67.7\pm0.5$\,{\scriptsize\negdelta{$(-1.2)$}}
    & $67.0\pm0.9$\,{\scriptsize\negdelta{$(-1.9)$}}
    & $66.7\pm1.3$\,{\scriptsize\negdelta{$(-2.2)$}}
    \\

    \rowcolor{blue!10}
    \quad\text{- }\textsc{\method}
    & $65.0\pm2.2$
    & $69.5\pm3.3$\,{\scriptsize\bfseries\posdelta{$(+4.5)$}}
    & $72.9\pm3.7$\,{\scriptsize\bfseries\posdelta{$(+7.9)$}}
    & $73.1\pm2.8$\,{\scriptsize\bfseries\posdelta{$(+8.1)$}}
    & $73.5\pm3.4$\,{\scriptsize\bfseries\posdelta{$(+8.5)$}}
    \\

    \rowcolor{blue!10}
    \quad\text{- }\textbf{Strict-OOD \textsc{\method}}
    & $62.7\pm1.4$
    & $66.8\pm5.0$\,{\scriptsize\bfseries\posdelta{$(+4.1)$}}
    & $66.2\pm4.0$\,{\scriptsize\bfseries\posdelta{$(+3.5)$}}
    & $72.5\pm3.0$\,{\scriptsize\bfseries\posdelta{$(+9.8)$}}
    & $72.0\pm0.8$\,{\scriptsize\bfseries\posdelta{$(+9.3)$}}
    \\
    \bottomrule
    \end{tabular}
    }
\end{table}

\begin{wraptable}{r}{0.3\textwidth}
    \vspace{-5pt}
    \centering
    \caption{Model performance on SkillsBench. Values are averaged across three runs.}
    \label{tab:skillsbench}
    \small
    \setlength{\tabcolsep}{4pt}
    \renewcommand{\arraystretch}{1.05}
    \scalebox{0.78}{
    \begin{tabular}{lll}
    \toprule
    Model & Score (\%) & Pass (\%) \\
    \midrule
    GLM-5.2 & 65.1 & 74.7 \\
    \hdashline
    Baseline & 33.3 & 47.1 \\
    \textbf\method & \textbf{41.4 {\scriptsize\posdelta{$(+8.1)$}}} & \textbf{62.1 {\scriptsize\posdelta{$(+15.0)$}}} \\
    \bottomrule
    \end{tabular}
    }
    \vspace{-6pt}
    \end{wraptable}
\paragraph{SkillsBench.}

Unlike the preceding benchmarks, SkillsBench holds the experience itself fixed: every model receives the same expert-curated skill, allowing us to isolate how effectively provided experience is used during execution. As shown in Table~\ref{tab:skillsbench}, \method improves over its matched backbone from $33.3$ to $41.4$ in average score and from $47.1\%$ to $62.1\%$ in average pass rate. These results provide complementary evidence that \method improves not only experience acquisition but also the ability to translate available experience into successful task execution.

\subsection{Domain-Excluded OOD Evaluation}
\label{sec:strict-ood}

Although the main evaluation benchmarks are excluded from training, the scaled training corpus may still contain semantically related domains. To impose a stricter distribution shift, we remove Software Engineering \& Programming, Finance \& Business, and IT Software \& Web Services from the training data. We apply a strict-prefix filter: a trajectory is excluded if either its current task or any preceding task in the same family belongs to one of these domains, preventing retained trajectories from depending on earlier in-domain experience. This leaves 3,634 trajectories, including 1,939 Shared-Tool and 1,695 Shared-Workflow trajectories. Starting from the same Qwen3.5-122B-A10B checkpoint, we train a Strict-OOD variant with the same configuration and evaluate its epoch-2 checkpoint on the same SpreadsheetBench splits under \protocol.

Table~\ref{tab:strict-ood} shows that the experience-learning gains persist even after removing training domains most closely related to SpreadsheetBench. Strict-OOD \method reaches $\Delta_{20}=+9.8$ and $\Delta_{40}=+9.3$, comparable to the gains of the full \method model. Despite its lower zero-experience performance, it exhibits substantial positive evolution as experience accumulates, whereas the matched untrained backbone consistently degrades under the same \protocol setting. This contrast provides stronger evidence that the learned experience-learning behavior transfers across domains and does not depend on training examples from related task domains.

\begin{table}[!t]
    \caption{Ablation of cross-family task designs on models. Values are mean $\pm$ population standard deviation over three runs, with $\Delta=S_{\mathrm{final}}-S_{\mathrm{initial}}$ shown in parentheses.}
    \label{tab:ablation}
    \centering
    \small
    \setlength{\tabcolsep}{3.5pt}
    \renewcommand{\arraystretch}{1.12}
    \resizebox{\linewidth}{!}{%
    \begin{tabular}{lcccccc}
    \toprule
    \multirow{2}{*}{\textbf{Benchmark}}
    & \multicolumn{2}{c}{\textbf{Shared-Tool Only}}
    & \multicolumn{2}{c}{\textbf{Shared-Workflow Only}}
    & \multicolumn{2}{c}{\textbf{Full \method}} \\
    \cmidrule(lr){2-3}\cmidrule(lr){4-5}\cmidrule(lr){6-7}
    & \textbf{Initial} & \textbf{Final} & \textbf{Initial} & \textbf{Final} & \textbf{Initial} & \textbf{Final} \\
    \midrule
    
    \rowcolor{black!8}
    \multicolumn{7}{c}{\textit{Task success rate} (\%, $\uparrow$)} \\
    
    SpreadsheetBench
    & $62.7\pm1.0$ & $72.4\pm7.3$\,{\scriptsize\posdelta{$(+9.7)$}}
    & $63.6\pm1.5$ & $69.7\pm2.7$\,{\scriptsize\posdelta{$(+6.1)$}}
    & $65.0\pm2.2$ & $73.5\pm3.4$\,{\scriptsize\posdelta{$(+8.5)$}} \\
    
    APEX-Agents
    & $13.2\pm1.3$ & $12.7\pm1.3$\,{\scriptsize\negdelta{$(-0.5)$}}
    & $12.2\pm0.6$ & $12.6\pm0.7$\,{\scriptsize\posdelta{$(+0.4)$}}
    & $12.2\pm1.7$ & $14.0\pm1.4$\,{\scriptsize\posdelta{$(+1.8)$}} \\
    
    AutomationBench
    & $7.2\pm0.5$ & $14.8\pm3.8$\,{\scriptsize\posdelta{$(+7.6)$}}
    & $8.5\pm0.5$ & $15.6\pm3.6$\,{\scriptsize\posdelta{$(+7.1)$}}
    & $14.5\pm0.9$ & $29.3\pm3.3$\,{\scriptsize\posdelta{$(+14.8)$}} \\
    
    \midrule
    
    \rowcolor{black!8}
    \multicolumn{7}{c}{\textit{Execution turns} ($\downarrow$)} \\
    
    SpreadsheetBench
    & $7.2\pm0.3$ & $10.5\pm1.7$\,{\scriptsize\negdelta{$(+3.3)$}}
    & $7.8\pm0.2$ & $9.0\pm0.4$\,{\scriptsize\negdelta{$(+1.2)$}}
    & $8.1\pm0.1$ & $8.6\pm0.7$\,{\scriptsize\negdelta{$(+0.5)$}} \\
    
    APEX-Agents
    & $29.0\pm0.7$ & $26.2\pm0.4$\,{\scriptsize\posdelta{$(-2.8)$}}
    & $31.2\pm0.5$ & $28.6\pm0.2$\,{\scriptsize\posdelta{$(-2.6)$}}
    & $33.3\pm0.6$ & $33.1\pm1.6$\,{\scriptsize\posdelta{$(-0.2)$}} \\
    
    AutomationBench
    & $43.0\pm0.7$ & $44.5\pm6.4$\,{\scriptsize\negdelta{$(+1.5)$}}
    & $45.7\pm1.2$ & $43.3\pm5.2$\,{\scriptsize\posdelta{$(-2.4)$}}
    & $49.4\pm1.1$ & $42.4\pm3.1$\,{\scriptsize\posdelta{$(-7.0)$}} \\
    
    \bottomrule
    \end{tabular}%
    }
    \end{table}

\subsection{Ablation Study}
\label{sec:ablation}

We ablate the two cross-task family designs by training separate models on Shared-Tool or Shared-Workflow trajectories, starting from the same base checkpoint. Table~\ref{tab:ablation} shows that both designs can induce experience-driven improvement, but with different behavioral patterns. Shared-Tool yields the larger gain on SpreadsheetBench ($+9.7$ vs.\ $+6.1$ points), consistent with learning reusable operational experience over stable black-box tools. Shared-Workflow, in contrast, more consistently improves execution efficiency on long-horizon tasks, reducing execution by $2.6$ turns on APEX-Agents and $2.4$ turns on AutomationBench, consistent with learning higher-level procedural experience across changing tool realizations. Combining both designs yields the broadest improvement across benchmarks. On AutomationBench in particular, full \method reaches a $+14.8$-point gain, substantially exceeding either Shared-Tool ($+7.6$) or Shared-Workflow ($+7.1$) alone, while also reducing execution by $7.0$ turns. These results suggest that the two family designs provide complementary supervision for transferable experience learning.

\subsection{Behavioral Analysis}

To better understand the source of the experience-learning gains, we analyze how accumulated skills are used and updated on SpreadsheetBench. \method-Qwen3.5 invokes existing skills far more frequently than its matched backbone despite maintaining a smaller bank, and its positive performance changes are concentrated on trajectories where skills are actually invoked. We further find that \method-Qwen3.5 consolidates feedback into fewer, longer skills with more failure- and verification-oriented content and less instance-specific structure. These patterns are consistent with the two behaviors targeted during training. See full analysis in Appendix~\ref{sec:further-analysis}.

\section{Conclusion}
\label{sec:conclusion}

We study whether experience learning itself can be trained as a transferable model capability. \method uses verifiable cross-task families to train models to acquire, update, and reuse external experience, while leaving model parameters fixed during downstream evolution. Across heterogeneous held-out benchmarks, the trained model consistently benefits more from accumulated experience than its matched backbone, with the effect persisting under domain-excluded training and extending to the use of externally provided skills. These results suggest that experience learning can be learned and transferred beyond the task distributions used for training.

\subsection*{Acknowledgments}
Min Yang is supported by National Key Research and Development Program of China (2024YFF0908200), National Natural Science Foundation of China (62376262), the Natural Science Foundation of Guangdong Province of China (2024A1515030166, 2025B1515020032).
This work was supported in part by the Science and Technology Development Fund of Macau SAR (Grant Nos. FDCT/0007/2024/AKP, EF2024-00185-FST), the UM and UMDF (Grant Nos. MYRG-GRG2024-00165-FST-UMDF, MYRG-GRG2025-00236-FST, MYRG-GRG2026-00018-FIC, MYRG-CRG2026-00007-FIC), and the Dr. Stanley Ho Medical Development Foundation (Grant No. SHMDF-AI/2026/001).

This work was supported by Alibaba Research Intern Program. We would like to thank the Qwen Team at Alibaba Token Hub (ATH), Alibaba Group, for providing the computational resources and foundation models (Qwen) used in this research.

\bibliography{ref}

@inproceedings{yao2023react,
  title={{ReAct}: Synergizing Reasoning and Acting in Language Models},
  author={Yao, Shunyu and Zhao, Jeffrey and Yu, Dian and Du, Nan and Shafran, Izhak and Narasimhan, Karthik and Cao, Yuan},
  booktitle={International Conference on Learning Representations},
  year={2023}
}

@misc{anthropic2025equipping,
  author={Anthropic},
  title={Equipping Agents for the Real World with Agent Skills},
  howpublished={https://www.anthropic.com/engineering/equipping-agents-for-the-real-world-with-agent-skills},
  year={2025}
}

@article{chen2026connect,
  title={Connect the Dots: Training LLMs for Long-Lifecycle Agents with Cross-Domain Generalization Via Reinforcement Learning},
  author={Chen, Yanxi and Shi, Weijie and Hu, Boyi and Zhang, Zeyue and Wang, Zhiwei and Xie, Yuexiang and Li, Yaliang and Ding, Bolin and Zhou, Jingren},
  journal={arXiv preprint arXiv:2606.20002},
  year={2026}
}

@inproceedings{shinn2023reflexion,
  title={Reflexion: Language Agents with Verbal Reinforcement Learning},
  author={Shinn, Noah and Cassano, Federico and Gopinath, Ashwin and Narasimhan, Karthik and Yao, Shunyu},
  booktitle={Advances in Neural Information Processing Systems},
  year={2023}
}

@inproceedings{zhao2024expel,
  title={{ExpeL}: {LLM} Agents Are Experiential Learners},
  author={Zhao, Andrew and Huang, Daniel and Xu, Quentin and Lin, Matthieu and Liu, Yong-Jin and Huang, Gao},
  booktitle={Proceedings of the AAAI Conference on Artificial Intelligence},
  volume={38},
  pages={19632--19642},
  year={2024}
}

@article{wang2024voyager,
  title={Voyager: An Open-Ended Embodied Agent with Large Language Models},
  author={Guanzhi Wang and Yuqi Xie and Yunfan Jiang and Ajay Mandlekar and Chaowei Xiao and Yuke Zhu and Linxi Fan and Anima Anandkumar},
  journal={Transactions on Machine Learning Research},
  issn={2835-8856},
  year={2024},
}

@article{ni2026trace2skill,
  title={{Trace2Skill}: Distill trajectory-local lessons into transferable agent skills},
  author={Ni, Jingwei and Liu, Yihao and Liu, Xinpeng and Sun, Yutao and Zhou, Mengyu and Cheng, Pengyu and Wang, Dexin and Zhao, Erchao and Jiang, Xiaoxi and Jiang, Guanjun},
  journal={arXiv preprint arXiv:2603.25158},
  year={2026}
}

@article{alzubi2026evoskill,
  title={{EvoSkill}: Automated skill discovery for multi-agent systems},
  author={Alzubi, Salaheddin and Provenzano, Noah and Bingham, Jaydon and Chen, Weiyuan and Vu, Tu},
  journal={arXiv preprint arXiv:2603.02766},
  year={2026}
}

@article{qiu2026autorefine,
  title={{AutoRefine}: From trajectories to reusable expertise for continual llm agent refinement},
  author={Qiu, Libin and Gao, Zhirong and Chen, Junfu and Ye, Yuhang and Huang, Weizhi and Xue, Xiaobo and Qiu, Wenkai and Tang, Shuo},
  journal={arXiv preprint arXiv:2601.22758},
  year={2026}
}

@article{jiang2026xskill,
  title={{XSkill}: Continual learning from experience and skills in multimodal agents},
  author={Jiang, Guanyu and Su, Zhaochen and Qu, Xiaoye and Fung, Yi R},
  journal={arXiv preprint arXiv:2603.12056},
  year={2026}
}

@article{lin2026muse,
  title={{Muse-AutoSkill}: Self-evolving agents via skill creation, memory, management, and evaluation},
  author={Lin, Huawei and Li, Peng and Song, Jie and Jiang, Fuxin and Zhang, Tieying},
  journal={arXiv preprint arXiv:2605.27366},
  year={2026}
}

@article{yu2026skilladaptor,
  title={{SkillAdaptor}: Self-adapting skills for llm agents from trajectories},
  author={Yu, Zhuoyun and Xie, Xin and Yao, Wuguannan and Wang, Chenxi and Liang, Lei and Qi, Xiang and Deng, Shumin},
  journal={arXiv preprint arXiv:2606.01311},
  year={2026}
}

@article{wang2026skillgrad,
  title={{SkillGrad}: Optimizing Agent Skills Like Gradient Descent},
  author={Wang, Hanyu and Lan, Yifan and Cao, Bochuan and Lin, Lu and Chen, Jinghui},
  journal={arXiv preprint arXiv:2605.27760},
  year={2026}
}

@article{yang2026skillopt,
  title={{SkillOpt}: Executive Strategy for Self-Evolving Agent Skills},
  author={Yang, Yifan and Gong, Ziyang and Huang, Weiquan and Yang, Qihao and Zhou, Ziwei and Huang, Zisu and Li, Yan and Gao, Xuemei and Dai, Qi and Liu, Bei and Qiu, Kai and Yang, Yuqing and Chen, Dongdong and Yang, Xue and Luo, Chong},
  journal={arXiv preprint arXiv:2605.23904},
  year={2026}
}

@article{li2026mind,
  title={MIND-Skill: Quality-Guaranteed Skill Generation via Multi-Agent Induction and Deduction},
  author={Li, Yixuan and Cai, Mingshu and Xiao, Ziyang and Wang, Wanyuan and Deng, Yanchen and An, Bo},
  journal={arXiv preprint arXiv:2605.08670},
  year={2026}
}

@inproceedings{wang2025agent,
title={Agent Workflow Memory},
author={Zora Zhiruo Wang and Jiayuan Mao and Daniel Fried and Graham Neubig},
booktitle={Forty-second International Conference on Machine Learning},
year={2025},
}

@inproceedings{wang2025inducing,
title={Inducing Programmatic Skills for Agentic Tasks},
author={Zora Zhiruo Wang and Apurva Gandhi and Graham Neubig and Daniel Fried},
booktitle={Second Conference on Language Modeling},
year={2025},
}

@inproceedings{zhang2026coevoskills,
title={{CoEvoSkills}: Self-Evolving Agent Skills via Co-Evolutionary Verification},
author={Hanrong Zhang and Shicheng Fan and Henry Peng Zou and Yankai Chen and Zhenting Wang and Jiayu Zhou and Chengze Li and Wei-Chieh Huang and Yifei Yao and Kening Zheng and Xue Liu and Xiaoxiao Li and Philip S. Yu},
booktitle={Third Conference on Language Modeling},
year={2026},
}

@article{liu2026skillrevise,
  title={{SkillRevise}: Improving LLM-Authored Agent Skills via Trace-Conditioned Skill Revision},
  author={Liu, Yuxuan and Su, Zhaochen and Xie, Lingyun and Zhang, Yuhao and Zong, Qing and Guo, Jiahe and Xie, Zhongwei and Ji, Yiyan and Yim, Yauwai and Luo, Hongyu and others},
  journal={arXiv preprint arXiv:2606.01139},
  year={2026}
}

@article{lu2026skill0,
  title={Skill0: In-context agentic reinforcement learning for skill internalization},
  author={Lu, Zhengxi and Yao, Zhiyuan and Wu, Jinyang and Han, Chengcheng and Gu, Qi and Cai, Xunliang and Lu, Weiming and Xiao, Jun and Zhuang, Yueting and Shen, Yongliang},
  journal={arXiv preprint arXiv:2604.02268},
  year={2026}
}

@article{zhu2026skill0,
  title={Skill0.5: Joint Skill Internalization and Utilization for Out-of-Distribution Generalization in Agentic Reinforcement Learning},
  author={Zhu, Jiapeng and Yu, Jianxiang and Zhao, Yibo and Han, Chengcheng and Gu, Qi and Cai, Xunliang and Li, Xiang and Qian, Weining},
  journal={arXiv preprint arXiv:2605.28424},
  year={2026}
}

@inproceedings{wang-etal-2026-reinforcement,
    title = "Reinforcement Learning for Self-Improving Agent with Skill Library",
    author = "Wang, Jiongxiao  and
      Yan, Qiaojing  and
      Wang, Yawei  and
      Tian, Yijun  and
      Mishra, Soumya Smruti  and
      Xu, Zhichao  and
      Gandhi, Megha  and
      Xu, Panpan  and
      Cheong, Lin Lee",
    booktitle = "Proceedings of the 64th Annual Meeting of the {A}ssociation for {C}omputational {L}inguistics (Volume 1: Long Papers)",
    year = "2026",
    pages = "1529--1550",
}

@article{wu2025evolver,
  title={{EvolveR}: Self-evolving llm agents through an experience-driven lifecycle},
  author={Wu, Rong and Wang, Xiaoman and Mei, Jianbiao and Cai, Pinlong and Fu, Daocheng and Yang, Cheng and Wen, Licheng and Yang, Xuemeng and Shen, Yufan and Wang, Yuxin and others},
  journal={arXiv preprint arXiv:2510.16079},
  year={2025}
}

@article{xia2026skillrl,
  title={{SkillRL}: Evolving Agents via Recursive Skill-Augmented Reinforcement Learning},
  author={Xia, Peng and Chen, Jianwen and Wang, Hanyang and Liu, Jiaqi and Zeng, Kaide and Wang, Yu and Han, Siwei and Zhou, Yiyang and Zhao, Xujiang and Chen, Haifeng and Zheng, Zeyu and Xie, Cihang and Yao, Huaxiu},
  journal={arXiv preprint arXiv:2602.08234},
  year={2026}
}

@article{yang2026skillmaster,
  title={{SkillMaster}: Toward Autonomous Skill Mastery in LLM Agents},
  author={Yang, Min and Piao, Jinghua and Xia, Xu and Lan, Xiaochong and Chen, Jiaju and Gong, Yongshun and Li, Yong},
  journal={arXiv preprint arXiv:2605.08693},
  year={2026}
}

@article{shi2026skill1,
  title={Skill1: Unified evolution of skill-augmented agents via reinforcement learning},
  author={Shi, Yaorui and Chen, Yuxin and Lu, Zhengxi and Miao, Yuchun and Liu, Shugui and Gu, Qi and Cai, Xunliang and Wang, Xiang and Zhang, An},
  journal={arXiv preprint arXiv:2605.06130},
  year={2026}
}

@article{he2026reskill,
  title={ReSkill: Reconciling Skill Creation with Policy Optimization in Agentic RL},
  author={He, Zelin and Lin, Haotian and Han, Boran and Zhu, Wei and Fang, Haoyang and Wang, Bernie and Zhu, Xuan and Li, Runze and Reimherr, Matthew},
  journal={arXiv preprint arXiv:2606.01619},
  year={2026}
}

@article{zhang2026skillcomposer,
  title={SkillComposer: Learning to Evolve Agent Skills for Specification and Generalization},
  author={Zhang, Qi and Feng, Zhaopeng and Shi, Xiaonan and Hu, Xiaomeng and Liu, Chu and Xie, Pengjun and Wang, Xiaobin and Ye, Jieping and Hooi, Bryan and Wang, Haobo and others},
  journal={arXiv preprint arXiv:2606.06079},
  year={2026}
}

@article{yao2026skillrise,
  title={{SkillRise}: Agentic Reinforcement Learning for Cross-Task Skill Evolution},
  author={Yao, Zhiyuan and Chen, Yuxin and Lu, Zhengxi and Xu, Zishan and Sun, Yueqing and Guo, Yifu and Lu, Yuquan and Cai, Zhengzhou and Zhang, Kangning and Han, Zhuowen and Wang, Zi-Han and Ye, Ziang and Gu, Qi and Cai, Xunliang and Liu, Weiwen and Shen, Yongliang},
  journal={arXiv preprint arXiv:2607.26784},
  year={2026}
}

@article{li2026codeskill,
  title={{CodeSkill}: Learning Self-Evolving Skills for Coding Agents},
  author={Li, Yanzhou and Zhang, Yiran and Zhang, Xiaoyu and Liu, Xiaoxia and Liu, Yang},
  journal={arXiv preprint arXiv:2605.25430},
  year={2026}
}

@article{vishe2026skill,
  title={{Skill-R1}: Agent skill evolution via reinforcement learning},
  author={Vishe, Yash and Surana, Rohan and Jiang, Xunyi and Huang, Zihan and Li, Xintong and Kuang, Nikki Lijing and Yu, Tong and Rossi, Ryan A and Shang, Jingbo and McAuley, Julian and others},
  journal={arXiv preprint arXiv:2605.09359},
  year={2026}
}

@article{fan2026evolving,
  title={Evolving-RL: End-to-End Optimization of Experience-Driven Self-Evolving Capability within Agents},
  author={Fan, Zhiyuan and Jin, Wenwei and Zhang, Feng and Li, Bin and Dong, Yihong and Hu, Yao and Li, Jiawei},
  journal={arXiv preprint arXiv:2605.10663},
  year={2026}
}

@inproceedings{zhou2025proposeragentevaluator,
title={Proposer-Agent-Evaluator ({PAE}): Autonomous Skill Discovery For Foundation Model Internet Agents},
author={Yifei Zhou and Qianlan Yang and Kaixiang Lin and Min Bai and Xiong Zhou and Yu-Xiong Wang and Sergey Levine and Li Erran Li},
booktitle={Forty-second International Conference on Machine Learning},
year={2025},
}

@article{zheng2025skillweaver,
  title={{SkillWeaver}: Web agents can self-improve by discovering and honing skills},
  author={Zheng, Boyuan and Fatemi, Michael Y and Jin, Xiaolong and Wang, Zora Zhiruo and Gandhi, Apurva and Song, Yueqi and Gu, Yu and Srinivasa, Jayanth and Liu, Gaowen and Neubig, Graham and others},
  journal={arXiv preprint arXiv:2504.07079},
  year={2025}
}

@article{tan2026skt,
  title={{SKT}: Skill-Use Training at Scale via Verified Synthetic Data Generation},
  author={Tan, Zelin and Zhang, Yiqun and Li, Hao and Cui, Zhiyao and Geng, Hejia and Zhang, Shao and Zhang, Hangfan and Chen, Yang and Wang, Xiaosong and Wang, Lilong and Yin, Zhenfei and Hu, Shuyue and Zhang, Chen and Bai, Lei},
  journal={arXiv preprint arXiv:2608.02287},
  year={2026}
}

@article{wei2026steps,
  title={Towards Compositional Generalization of {LLM}s via Skill Taxonomy Guided Data Synthesis},
  author={Wei, Yifan and Du, Li and Yu, Xiaoyan and Feng, Yang and Li, Angsheng},
  journal={arXiv preprint arXiv:2601.03676},
  year={2026}
}

@inproceedings{xie2026agentsynth,
  title={{AgentSynth}: Scalable Task Generation for Generalist Computer-Use Agents},
  author={Xie, Jingxu and Xu, Dylan and Zhao, Xuandong and Song, Dawn},
  booktitle={The Fourteenth International Conference on Learning Representations},
  year={2026}
}

@article{zhu2026termigen,
  title={{TermiGen}: High-Fidelity Environment and Robust Trajectory Synthesis for Terminal Agents},
  author={Zhu, Kaijie and Nie, Yuzhou and Li, Yijiang and Huang, Yiming and Wu, Jialian and Liu, Jiang and Sun, Ximeng and Yin, Zhenfei and Wang, Lun and Liu, Zicheng and Barsoum, Emad and Wang, William Yang and Guo, Wenbo},
  journal={arXiv preprint arXiv:2602.07274},
  year={2026}
}

@article{hua2026cliuniverse,
  title={{CLI-Universe}: Towards Verifiable Task Synthesis Engine for Terminal Agents},
  author={Hua, Zhanbo and Yao, Yifan and Xie, Weihao and Zhao, Yongchi and Liu, Minghao and Qiu, Ruizhi and Huang, Zhewei and Wang, Zun and Ji, Yiyan and Ye, Yunhai and Zhu, Letian and Lei, Xinping and Li, Han and Ma, Zhiyuan and Wang, Zili and Zhang, Zhaoxiang and Liu, Jiaheng},
  journal={arXiv preprint arXiv:2606.22883},
  year={2026}
}

@article{fan2026skillsynth,
  title={Toward Scalable Terminal Task Synthesis via Skill Graphs},
  author={Fan, Zhiyuan and Yu, Tinghao and Cai, Yuanjun and Guan, Jiangtao and Yang, Yun and Hu, Dingxin and Zhou, Jiang and Wu, Xing and Han, Zhuo and Zhang, Feng and Wang, Lilin},
  journal={arXiv preprint arXiv:2604.25727},
  year={2026}
}

@article{cheng2026terminalworld,
  title={{Terminal-World}: Scaling Terminal-Agent Environments via Agent Skills},
  author={Cheng, Zihao and Wang, Hongru and Liu, Zeming and Wang, Xinyi and Zhu, Xiangrong and Guo, Yuhang and Lin, Wei and Pan, Jeff Z. and Wang, Yunhong},
  journal={arXiv preprint arXiv:2605.20876},
  year={2026}
}

@article{li2026skillsbench,
  title={{SkillsBench}: Benchmarking How Well Agent Skills Work Across Diverse Tasks},
  author={Li, Xiangyi and Liu, Yimin and Chen, Wenbo and You, Bingran and Di, Zonglin and He, Yifeng and Zheng, Shenghan and Choe, Kyoung Whan and Sun, Jiankai and Wang, Shuyi and Tao, Chujun and Li, Binxu and Zhao, Xuandong and Geng, Hejia and Wu, Xiaojun and Zhou, Junwei and Chen, Xiaokun and Xing, Hanwen and Li, Yubo and Zeng, Qunhong and Wang, Di and Wang, Yuanli and Chaim, Roey Ben and Jiang, Penghao and Shen, Haotian and Kong, Luyang and Liu, Xinyi and Wang, Runhui and Liu, Xuanqing and Li, Jiachen and Lan, Xin and Lin, Yueqian and Ye, Wengao and He, Junwei and Li, Songlin and Zhang, Yue and Gao, Yipeng and Li, Yijiang and Ma, Ze and Jing, Liqiang and Wang, Tianyu and Li, Kaixin and Xue, Yiqi and Lyu, Haoran and He, Yizhuo and Tian, Yuchen and Wu, Shutong and Wang, Bowei and Gao, Yixuan and Chen, Bo and Liu, Litong and Cheng, Sikai and Bao, Jiajun and Tong, Shuaicheng and Xu, Shuwen and Zhuo, Terry Yue and Ye, Tinghan and Qi, Qi and Li, Miao and Liao, Longtai and Tan, Zelin and Shi, Chang and Tang, Xilin and Tankasala, Srinath and Yuan, Boqin and Qian, Yaoyao and Tu, Jianhong and Wang, Chenguang and Sun, Yizhou and Wang, Wei and Taylor, Aaron and Yang, Ziyue and Guan, Changkun and Dong, Zhikang and Zhang, Xinyu and Dillmann, Steven and Lee, Han-chung and Song, Dawn},
  journal={arXiv preprint arXiv:2602.12670},
  year={2026}
}

@article{han2026swe,
  title={{SWE-Skills-Bench}: Do Agent Skills Actually Help in Real-World Software Engineering?},
  author={Han, Tingxu and Zhang, Yi and Song, Wei and Fang, Chunrong and Chen, Zhenyu and Sun, Youcheng and Hu, Lijie},
  journal={arXiv preprint arXiv:2603.15401},
  year={2026}
}

@article{zhong2026skilllearnbench,
  title={{SkillLearnBench}: Benchmarking Continual Learning Methods for Agent Skill Generation on Real-World Tasks},
  author={Zhong, Shanshan and Lu, Yi and Ning, Jingjie and Wan, Yibing and Feng, Lihan and Ao, Yuyi and Ribeiro, Leonardo FR and Dreyer, Markus and Ammirati, Sean and Xiong, Chenyan},
  journal={arXiv preprint arXiv:2604.20087},
  year={2026}
}

@article{zhang2026skillflow,
  title={{SkillFlow}: Benchmarking Lifelong Skill Discovery and Evolution for Autonomous Agents},
  author={Zhang, Ziao and Shi, Kou and Huang, Shiting and Nie, Avery and Zeng, Yu and Zhao, Yiming and Fang, Zhen and Su, Qishen and Qiu, Haibo and Yang, Wei and others},
  journal={arXiv preprint arXiv:2604.17308},
  year={2026}
}

@article{zhou2026skillgenbench,
  title={{SkillGenBench}: Benchmarking skill generation pipelines for llm agents},
  author={Zhou, Yifan and Zhang, Zhentao and Cheng, Ziming and Zhang, Shuo and Lan, Qizhen and Chen, Zhangquan and Yang, Zhi and Chen, Ronghao and Wang, Huacan and Hu, Sen and others},
  journal={arXiv preprint arXiv:2605.18693},
  year={2026}
}

@article{lei2026skillevolbench,
  title={{SkillEvolBench}: Benchmarking the Evolution from Episodic Experience to Procedural Skills},
  author={Lei, Yingtie and Wan, Zhongwei and Zhang, Jiankun and Alam, Samiul and Zhong, Zixuan and Huang, Peizhou and Wang, Xin and Zhang, Jingxuan and Zhou, Donghao and Hsieh, Yunta and others},
  journal={arXiv preprint arXiv:2605.24117},
  year={2026}
}

@article{jiang2026sea,
  title={{SEA-Eval}: A Benchmark for Evaluating Self-Evolving Agents Beyond Episodic Assessment},
  author={Jiang, Sihang and Ma, Lipeng and Hong, Zhonghua and Wang, Keyi and Lu, Zhiyu and Wang, Tengfei and Chen, Shisong and Zhang, Jinghao and Pan, Tianjun and Li, Weijia and others},
  journal={arXiv preprint arXiv:2604.08988},
  year={2026}
}

@article{zhou2026gdpevo,
  title={{GDPevo}: Evaluating Agent Self-Evolution on Real Business Tasks},
  author={Zhou, Leijun and Liu, Zhihao and Qu, Xiang and Liu, Chenxu and Liu, Yifei and Yu, Yanke and Xu, Jingzhe and Wu, Xuejun and Qian, Buyue and Chen, Xi and others},
  journal={arXiv preprint arXiv:2608.03764},
  year={2026}
}

@article{gao2026evoagentbench,
  title={{EvoAgentBench}: Benchmarking Agent Self-Evolution via Ability Transfer},
  author={Gao, Xingze and Hu, Chuanrui and Chen, Hongda and Yao, Pengfei and Wang, Zhao and Bai, Yi and Wu, Zhengwei and Han, Yunyun and Cong, Xiaofeng and Gui, Jie and others},
  journal={arXiv preprint arXiv:2607.05202},
  year={2026}
}

@misc{qwen2026qwen35,
    title  = {{Qwen3.5}: Towards Native Multimodal Agents},
    author = {{Qwen Team}},
    month  = {February},
    year   = {2026},
}

@article{zeng2026glm,
  title={{GLM-5}: From vibe coding to agentic engineering},
  author={Zeng, Aohan and Lv, Xin and Hou, Zhenyu and Du, Zhengxiao and Zheng, Qinkai and Chen, Bin and Yin, Da and Ge, Chendi and Huang, Chenghua and Xie, Chengxing and others},
  journal={arXiv preprint arXiv:2602.15763},
  year={2026}
}
\bibliographystyle{colm2024_conference}

\clearpage
\appendix

\begin{table}[!t]
    \caption{SpreadsheetBench experience-access behavior. Values are means $\pm$ population standard deviations over three seeds.}
    \label{tab:skill-invocation}
    \centering
    \scriptsize
    \begin{tabular}{llrrrr}
    \toprule
    Model & Condition & Skills in bank & Any invocation (\%) & Skill as first tool (\%) & Calls/task \\
    \midrule
    \multirow{4}{*}{Baseline-Qwen3.5}
    & $k=5$ & $3.0\pm0.0$ & $9.3\pm2.0$ & $2.4\pm1.3$ & $0.094\pm0.021$ \\
    & $k=10$ & $5.3\pm0.5$ & $11.3\pm2.5$ & $4.2\pm1.5$ & $0.113\pm0.025$ \\
    & $k=20$ & $10.0\pm0.0$ & $18.6\pm4.7$ & $7.8\pm0.1$ & $0.188\pm0.049$ \\
    & $k=40$ & $10.0\pm0.0$ & $17.0\pm8.2$ & $5.0\pm2.6$ & $0.172\pm0.084$ \\
    \midrule
    \multirow{4}{*}{\method-Qwen3.5}
    & $k=5$ & $2.0\pm0.8$ & $76.4\pm16.4$ & $63.1\pm23.2$ & $0.765\pm0.162$ \\
    & $k=10$ & $3.3\pm2.6$ & $79.5\pm16.9$ & $67.0\pm23.0$ & $0.800\pm0.164$ \\
    & $k=20$ & $4.3\pm4.0$ & $81.0\pm12.3$ & $67.8\pm15.4$ & $0.811\pm0.121$ \\
    & $k=40$ & $4.7\pm3.9$ & $81.5\pm10.0$ & $70.6\pm15.0$ & $0.819\pm0.094$ \\
    \bottomrule
    \end{tabular}
\end{table}
    
\begin{table}[!t]
    \caption{Paired success-rate change relative to the same task under an empty-bank condition (\%).}
    \label{tab:invocation-utility}
    \centering
    \small
    \begin{tabular}{llrr}
    \toprule
    Model & Conditions & Invoked skill & Did not invoke \\
    \midrule
    Baseline-Qwen3.5 & All $k$ pooled & $-4.2$ & $-1.2$ \\
    \method-Qwen3.5 & All $k$ pooled & $+9.5$ & $-1.9$ \\
    \method-Qwen3.5 & $k=20$ & $+10.8\pm4.7$ & $-0.3\pm4.8$ \\
    \method-Qwen3.5 & $k=40$ & $+10.2\pm6.4$ & $-4.7\pm6.7$ \\
    \bottomrule
    \end{tabular}
\end{table}    

\section{Further Analysis}
\label{sec:further-analysis}

The main results show that \method-Qwen3.5 benefits substantially more from accumulated experience than its matched backbone. We further examine how this difference manifests in the two behaviors targeted during training: using accumulated experience during task execution and updating that experience from feedback. Our analysis focuses on SpreadsheetBench and compares \method-Qwen3.5-122B-A10B with its matched Qwen3.5-122B-A10B backbone, referred to below as \method-Qwen3.5 and Baseline-Qwen3.5.

\subsection{Experience-Conditioned Execution}

We analyze 8,399 valid SpreadsheetBench evaluation trajectories across both models, three seeds, and $k\in\{5,10,20,40\}$. A skill invocation is counted only when the model calls a generated skill available in the current bank; built-in harness skills and invalid skill names are excluded. Table~\ref{tab:skill-invocation} summarizes how often and how early each model accesses its accumulated skills during execution. Across the four conditions, Baseline-Qwen3.5 invokes a skill on 14.1\% of tasks, compared with 79.6\% for \method-Qwen3.5. The difference is even larger for early use: a skill is the first tool invoked on 4.8\% of tasks for the baseline and 67.1\% for \method-Qwen3.5. This behavior cannot be attributed to a larger skill bank, since \method-Qwen3.5 maintains fewer skills throughout adaptation. Instead, the trained model consults its accumulated experience much more consistently and earlier in the execution process. 

More frequent invocation does not by itself indicate that the retrieved experience is beneficial. We therefore pair each task with the corresponding run from the same model and seed under an empty-bank condition. Table~\ref{tab:invocation-utility} reports the change in success rate relative to this matched run. For \method-Qwen3.5, tasks on which a skill is invoked improve by $+9.5$ points on average relative to their matched empty-bank runs. By contrast, the invoked subset of Baseline-Qwen3.5 performs $4.2$ points below its empty-bank counterpart. The gains of \method-Qwen3.5 are therefore closely associated with trajectories in which accumulated experience is actually used during execution.

\subsection{Feedback-Driven Experience Updating}

\begin{table}[!t]
\caption{Structure of the final SpreadsheetBench skill banks at $k=40$, pooled over three seeds.}
\label{tab:skill-structure}
\centering
\small
\begin{tabular}{lrr}
\toprule
Metric & Baseline-Qwen3.5 & \method-Qwen3.5 \\
\midrule
Skills & 30 & 14 \\
Lexical tokens & 29,679 & 31,155 \\
Tokens per skill & 989.3 & 2,225.4 \\
Failure terms / 1K tokens & 1.42 & 5.42 \\
Verification terms / 1K tokens & 3.13 & 4.56 \\
Opaque IDs / 1K tokens & 2.26 & 0.00 \\
Top-2 invocation share & 32.2\% & 74.5\% \\
\bottomrule
\end{tabular}
\end{table}

\begin{table}[!t]
\caption{Case study on how different models encode experiences on SpreadsheetBench.}
\label{tab:skill-cases}
\centering
\small
\begin{tabular}{p{2.2cm}p{4.6cm}p{5.3cm}}
\toprule
Case & Baseline-Qwen3.5 & \method-Qwen3.5 \\
\midrule
Cross-sheet lookup
& A local VLOOKUP/INDEX--MATCH recipe with several formula variants and a recommendation to write static values.
& A symptom--cause--repair--verification taxonomy covering truncated ranges, wrong-sheet references, hidden whitespace, cached formula values, and cell-level validation. \\
Boundary-sensitive editing
& A generic checklist: preserve values, use absolute references, handle missing keys, and verify the output.
& Conditional diagnostic rules for inflated worksheet bounds and locale-dependent formula separators, tied to observable failure signals and corrective checks. \\
\bottomrule
\end{tabular}
\end{table}

We next examine the skill banks produced through repeated feedback-driven updates. Because these banks reflect the model's decisions about what to retain, revise, or discard over adaptation, their final structure offers a complementary view of the learned experience-update behavior. Table~\ref{tab:skill-structure} compares the banks after 40 adaptation tasks. Although the two models produce similar total text volume, they organize that content quite differently. \method-Qwen3.5 stores substantially fewer skills, resulting in considerably longer skills on average. Its banks also contain more language related to failures and verification, no detected opaque identifiers, and a larger share of total invocations concentrated on a small number of skills. This pattern suggests stronger consolidation around reusable diagnostic and validation-oriented knowledge, with less fragmentation into narrowly scoped or instance-specific records.

Because greater length or concentration can also introduce risks such as overgeneralization, we emphasize that these statistics characterize bank structure rather than skill quality directly. Table~\ref{tab:skill-cases} provides qualitative examples of how these structural differences appear in the skill contents. Baseline-Qwen3.5 tends to record direct procedures or local recipes for completing specific operations. By contrast, \method-Qwen3.5 more often organizes experience around applicability conditions, possible causes, corrective actions, and subsequent verification. These examples align with the quantitative bank-level patterns above and suggest that meta-experience training encourages feedback to be consolidated into more structured and broadly reusable guidance, while reducing reliance on task-local procedures.

Taken together, the analyses indicate that meta-experience training changes both sides of the experience-learning process. \method-Qwen3.5 makes accumulated skills a more active part of subsequent task execution and organizes interaction feedback into a more consolidated form for future reuse.

\definecolor{sharedtool}{HTML}{6F8FAF}
\definecolor{sharedworkflow}{HTML}{C8906B}
\definecolor{seedbar}{HTML}{8496A8}
\definecolor{schemashare}{HTML}{8D83A6}
\definecolor{recordcoverage}{HTML}{8D969E}
\section{Seed Diversity for Environment Scaling}
\label{app:seed-distributions}

The family-scaling pipeline draws on heterogeneous seeds from pretraining-style corpora, task examples, and tool schemas. These sources provide complementary starting points for construction: corpus passages supply semantic contexts, task examples provide procedural patterns for workflow abstraction, and tool schemas provide executable affordances for tool-based interaction. We characterize their diversity below to illustrate the breadth of the substrate available for scaling controlled task families.

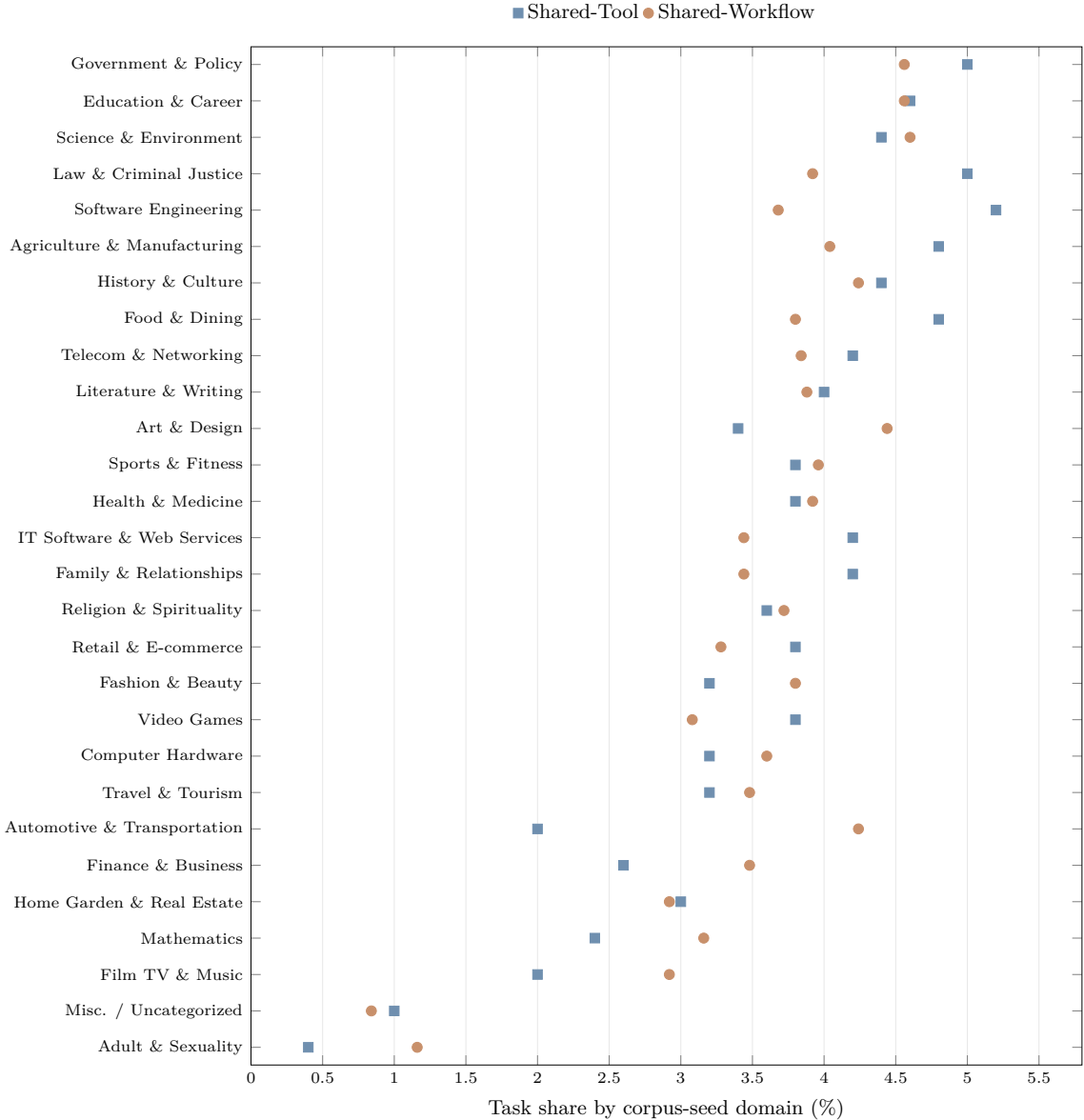
\begin{figure}[!t]
\centering
\scalebox{0.9}{
\begin{tikzpicture}
\begin{axis}[
    width=0.90\textwidth,
    height=0.70\textheight,
    xmin=0,
    xmax=5.8,
    xlabel={Task share by corpus-seed domain (\%)},
    symbolic y coords={
        Government \& Policy,
        Education \& Career,
        Science \& Environment,
        Law \& Criminal Justice,
        Software Engineering,
        Agriculture \& Manufacturing,
        History \& Culture,
        Food \& Dining,
        Telecom \& Networking,
        Literature \& Writing,
        Art \& Design,
        Sports \& Fitness,
        Health \& Medicine,
        IT Software \& Web Services,
        Family \& Relationships,
        Religion \& Spirituality,
        Retail \& E-commerce,
        Fashion \& Beauty,
        Video Games,
        Computer Hardware,
        Travel \& Tourism,
        Automotive \& Transportation,
        Finance \& Business,
        Home Garden \& Real Estate,
        Mathematics,
        Film TV \& Music,
        Misc. / Uncategorized,
        Adult \& Sexuality
    },
    ytick=data,
    y dir=reverse,
    tick label style={font=\scriptsize},
    label style={font=\small},
    legend style={
        font=\small,
        at={(0.5,1.015)},
        anchor=south,
        legend columns=2,
        draw=none
    },
    xmajorgrids=true,
    ymajorgrids=false,
    grid style={gray!18},
    enlarge y limits=0.018
]

\addplot[
    only marks,
    mark=square*,
    mark size=2.1pt,
    color=sharedtool
] coordinates {
    (5.00,Government \& Policy)
    (4.60,Education \& Career)
    (4.40,Science \& Environment)
    (5.00,Law \& Criminal Justice)
    (5.20,Software Engineering)
    (4.80,Agriculture \& Manufacturing)
    (4.40,History \& Culture)
    (4.80,Food \& Dining)
    (4.20,Telecom \& Networking)
    (4.00,Literature \& Writing)
    (3.40,Art \& Design)
    (3.80,Sports \& Fitness)
    (3.80,Health \& Medicine)
    (4.20,IT Software \& Web Services)
    (4.20,Family \& Relationships)
    (3.60,Religion \& Spirituality)
    (3.80,Retail \& E-commerce)
    (3.20,Fashion \& Beauty)
    (3.80,Video Games)
    (3.20,Computer Hardware)
    (3.20,Travel \& Tourism)
    (2.00,Automotive \& Transportation)
    (2.60,Finance \& Business)
    (3.00,Home Garden \& Real Estate)
    (2.40,Mathematics)
    (2.00,Film TV \& Music)
    (1.00,Misc. / Uncategorized)
    (0.40,Adult \& Sexuality)
};

\addplot[
    only marks,
    mark=*,
    mark size=2.2pt,
    color=sharedworkflow
] coordinates {
    (4.56,Government \& Policy)
    (4.56,Education \& Career)
    (4.60,Science \& Environment)
    (3.92,Law \& Criminal Justice)
    (3.68,Software Engineering)
    (4.04,Agriculture \& Manufacturing)
    (4.24,History \& Culture)
    (3.80,Food \& Dining)
    (3.84,Telecom \& Networking)
    (3.88,Literature \& Writing)
    (4.44,Art \& Design)
    (3.96,Sports \& Fitness)
    (3.92,Health \& Medicine)
    (3.44,IT Software \& Web Services)
    (3.44,Family \& Relationships)
    (3.72,Religion \& Spirituality)
    (3.28,Retail \& E-commerce)
    (3.80,Fashion \& Beauty)
    (3.08,Video Games)
    (3.60,Computer Hardware)
    (3.48,Travel \& Tourism)
    (4.24,Automotive \& Transportation)
    (3.48,Finance \& Business)
    (2.92,Home Garden \& Real Estate)
    (3.16,Mathematics)
    (2.92,Film TV \& Music)
    (0.84,Misc. / Uncategorized)
    (1.16,Adult \& Sexuality)
};

\legend{Shared-Tool,Shared-Workflow}

\end{axis}
\end{tikzpicture}
}
\caption{Domain distribution of corpus seeds under Shared-Tool and Shared-Workflow construction.}
\label{fig:corpus-seeds}
\end{figure}

\paragraph{Corpus seeds.}
Corpus passages provide the semantic context used during environment construction. Shared-Tool uses 500 independently sampled passages, each reused across the five members of one family, whereas Shared-Workflow independently samples passages for all 2,500 task instances. As shown in Figure~\ref{fig:corpus-seeds}, both construction modes cover all 28 domains, while independent sampling yields a slightly more even task-level distribution for Shared-Workflow.

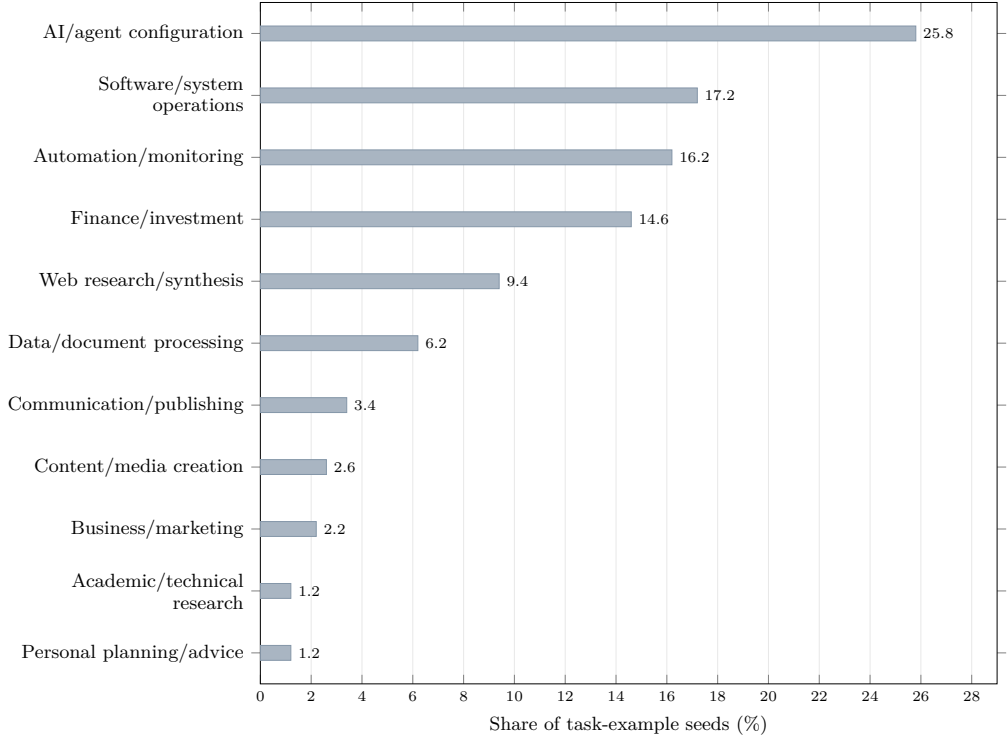
\begin{figure}[!t]
\centering
\scalebox{0.8}{
\begin{tikzpicture}
\begin{axis}[
    xbar,
    width=0.86\textwidth,
    height=0.52\textheight,
    xmin=0,
    xmax=29,
    xlabel={Share of task-example seeds (\%)},
    symbolic y coords={
        AI/agent configuration,
        Software/system operations,
        Automation/monitoring,
        Finance/investment,
        Web research/synthesis,
        Data/document processing,
        Communication/publishing,
        Content/media creation,
        Business/marketing,
        Academic/technical research,
        Personal planning/advice
    },
    ytick=data,
    y dir=reverse,
    yticklabel style={
        font=\small,
        text width=3.9cm,
        align=right
    },
    xticklabel style={font=\scriptsize},
    label style={font=\small},
    nodes near coords,
    nodes near coords style={font=\scriptsize},
    bar width=7pt,
    enlarge y limits=0.05,
    xmajorgrids=true,
    ymajorgrids=false,
    grid style={gray!18}
]

\addplot[
    fill=seedbar!70,
    draw=seedbar
] coordinates {
    (25.8,AI/agent configuration)
    (17.2,Software/system operations)
    (16.2,Automation/monitoring)
    (14.6,Finance/investment)
    (9.4,Web research/synthesis)
    (6.2,Data/document processing)
    (3.4,Communication/publishing)
    (2.6,Content/media creation)
    (2.2,Business/marketing)
    (1.2,Academic/technical research)
    (1.2,Personal planning/advice)
};

\end{axis}
\end{tikzpicture}
}
\caption{Primary-intent distribution of the task-example seeds used in \method.}
\label{fig:task-seeds}
\end{figure}

\paragraph{Task-example seeds.}
The 500 task examples provide candidate procedural patterns from which Shared-Workflow families can be abstracted. For analysis, each example is assigned one primary intent using a reproducible researcher annotation rather than a source-provided label. Figure~\ref{fig:task-seeds} shows that the seed set emphasizes agent configuration, software and system operations, automation, and finance, while retaining a long tail of other intents.

\begin{figure}[!t]
\centering
\scalebox{0.8}{
\begin{tikzpicture}
\begin{axis}[
    width=0.82\textwidth,
    height=0.40\textheight,
    xmin=0,
    xmax=72,
    xlabel={Tool-example share / seed-record coverage (\%)},
    symbolic y coords={
        Web information access,
        Code/system execution,
        File/content operations,
        Scheduling/automation,
        Agent/session orchestration,
        Visual canvas
    },
    ytick=data,
    y dir=reverse,
    yticklabel style={
        font=\small,
        text width=3.5cm,
        align=right
    },
    xticklabel style={font=\scriptsize},
    label style={font=\small},
    legend style={
        font=\small,
        at={(0.5,1.02)},
        anchor=south,
        legend columns=2,
        draw=none
    },
    xmajorgrids=true,
    ymajorgrids=false,
    grid style={gray!18},
    enlarge y limits=0.12
]

\addplot[
    only marks,
    mark=*,
    mark size=2.4pt,
    color=schemashare
] coordinates {
    (58.0,Web information access)
    (22.6,Code/system execution)
    (13.6,File/content operations)
    (2.7,Scheduling/automation)
    (2.6,Agent/session orchestration)
    (0.5,Visual canvas)
};

\addplot[
    only marks,
    mark=square*,
    mark size=2.3pt,
    color=recordcoverage
] coordinates {
    (66.4,Web information access)
    (39.4,Code/system execution)
    (34.2,File/content operations)
    (10.2,Scheduling/automation)
    (9.0,Agent/session orchestration)
    (2.4,Visual canvas)
};

\legend{Tool-schema share,Seed-record coverage}

\end{axis}
\end{tikzpicture}
}
\caption{Capability distribution of tool examples and their coverage across seed records.}
\label{fig:tool-seeds}
\end{figure}
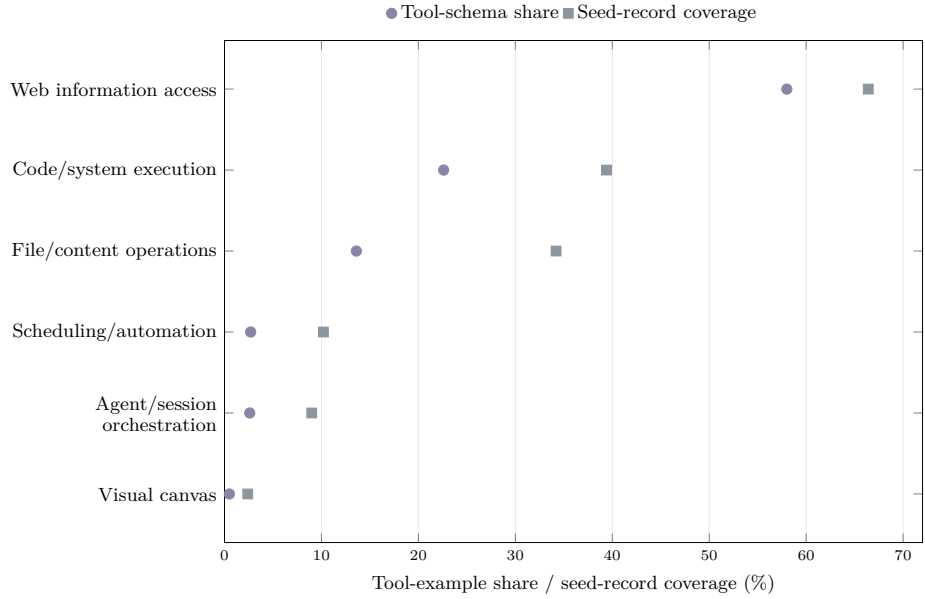

\paragraph{Tool-example seeds.}
The 500 seed records contain 2,338 tool-use examples, each pairing an interface schema with a corresponding usage example. These examples span 16 top-level tool names, which we deterministically group into six capability categories for analysis. Each seed record contains between 1 and 10 tool examples, with 4.676 examples on average. Figure~\ref{fig:tool-seeds} reports two complementary statistics. \emph{Tool-example share} measures the fraction of the 2,338 examples assigned to each capability category, while \emph{seed-record coverage} measures the fraction of the 500 seed records containing at least one example from that category.

Together, the three sources provide heterogeneous semantic contexts, procedural patterns, and executable affordances, giving the family-scaling pipeline diverse starting points without tying construction to a narrow set of manually specified environments. These statistics characterize the breadth of the scaling seeds rather than the final training distribution.

\section{Discussion}
\label{app:discussion}

Our current training strategy uses verified teacher trajectories to supervise experience-conditioned execution and feedback-driven experience updating. While this provides an efficient way to bootstrap transferable experience-learning behavior, it does not explicitly optimize the downstream utility of each experience update. A natural next question is therefore how future task performance can be incorporated more directly into training.

\paragraph{Why supervised fine-tuning?}

The environments, tasks, and executable verifiers constructed by \method readily support reinforcement learning. However, experience learning introduces a delayed cross-episode objective that is more difficult to optimize than task-level reward. An update to the skill bank may appear reasonable on the task that produced it, yet its true utility is only revealed when that experience is reused on subsequent tasks. Reinforcement learning would therefore require attributing future performance changes back to earlier execution and reflection decisions, creating a nontrivial credit-assignment problem. It would also require rolling out multiple downstream tasks to evaluate earlier updates, substantially increasing the cost of optimization over long sequential task streams.

For this reason, we use rejection sampling followed by supervised fine-tuning as a practical first step. Executable verification allows us to filter reliable teacher trajectories at substantially lower cost, providing the base model with an initial capability for acquiring and reusing experience before introducing more expensive cross-episode optimization.

\paragraph{Filtering by downstream experience utility.}

The task-family structure of \method also supports an intermediate approach between local trajectory filtering and full reinforcement learning. Because a family can contain many related but non-identical tasks, some members can be reserved as held-out validation tasks while the remaining members are used for sequential experience collection. After an update from $B_{i-1}$ to $B_i$, the two banks can be evaluated on the same held-out family members, yielding
\begin{equation}
    \Delta_i^{\mathrm{exp}}
    =
    S_{\mathrm{val}}(B_i)
    -
    S_{\mathrm{val}}(B_{i-1}).
\end{equation}

Unlike the task verifier, which measures whether the current task was solved correctly, $\Delta_i^{\mathrm{exp}}$ measures whether the resulting experience improves performance on related tasks that were not used to construct the bank. It can therefore provide a stronger rejection criterion for selecting teacher trajectories whose experience updates exhibit positive downstream transfer.

This strategy aligns trajectory selection more directly with the objective of experience learning, but requires additional evaluation rollouts for each candidate bank update and is consequently more expensive than the filtering used in this work. We leave downstream-utility-based filtering and reinforcement learning with cross-episode reward signals as promising directions for further improving meta-experience learning.

\section{Prompt Templates for Experience Learning}
\label{app:prompt-templates}

As described in Section~\ref{sec:method}, an agent trajectory consists of task execution followed by feedback-driven reflection. We use fixed harness-level templates for these two stages during both training-trajectory collection and experience-learning evaluation. The templates shown below are shared across models; runtime-provided context, such as the underlying agent system message, available-skill inventory, and execution reminders, are supplied separately by the specific agent harness.

\paragraph{Task execution.}
The execution template initializes each task, specifies the sandbox layout, and instructs the agent to consult relevant accumulated skills when available.

\begin{lstlisting}[
  basicstyle=\ttfamily\footnotesize,
  columns=fullflexible,
  breaklines=true,
  frame=single
]
# Goal
You are solving a task in a linux sandbox.
# Environment
- `/workspace/source/`: Data, tool packages, and README.md file
- `/workspace/input/`: Read-only task description and sub-task files
- `/workspace/output/`: Write ALL output files here (do not write elsewhere)
# Recommended Workflow
1. Read `/workspace/input/task.txt` to understand the task.
2. Explore `/workspace/source/` to understand available data.
3. If any available skill covers the task or a sub-step, invoke it first.
4. Write and execute bash/Python scripts to produce the required output.
5. Verify that all output files are in `/workspace/output/`.
6. Call `task_done` when finished.
\end{lstlisting}

\paragraph{Experience reflection.}
After task evaluation, feedback is injected into the same interaction session using the following template. The agent can therefore inspect its preceding execution trajectory together with the evaluation outcome before deciding whether to create, revise, or retain skills in the current bank.

\begin{lstlisting}[
  basicstyle=\ttfamily\footnotesize,
  columns=fullflexible,
  breaklines=true,
  frame=single
]
You are a reflective learning agent. Your goal is to convert task experience into reusable skills that improve future performance.
### Evaluation Result
Your performance on the previously assigned task has been evaluated. The reward score is:
<reward>
{reward}
</reward>
The score ranges from 0 (complete failure) to 1 (complete success).
<failing_rubrics>
{failing_rubrics}
</failing_rubrics>
### Objective
Based on this outcome, distill your experience into a reusable skill that will help you or another AI assistant perform better on similar tasks in the future. You may access and save skills to the following directory:
<skills_dir>
{skills_dir}
</skills_dir>
Each skill is a directory containing a SKILL.md file with YAML frontmatter (name, description) and markdown instructions. Use the skill-creator skill for guidance on creating or updating skills.
### Workflow
1. Analyze your task trajectory in light of the reward signal.
2. Check existing skills and decide whether to create a new one, update an existing one, or leave them unchanged.
3. Call `task_done` when complete.
### Rules
- Always use the skill-creator skill to create or update skills.
- DO NOT modify the `skill-creator` skill itself.
- The total number of available skills must not exceed **{max_skill_num}**. Only create a new skill when no existing skill adequately covers the scenario.
\end{lstlisting}

At runtime, \texttt{reward} contains the final task score, while \texttt{failing\_rubrics} contains either unsatisfied verifier criteria or error assertions, depending on the verifier type. \texttt{skills\_dir} specifies the mounted skill bank, and \texttt{max\_skill\_num} controls its capacity.

\end{document}